%% file: main.tex
\documentclass[10pt,twocolumn,letterpaper]{article}

\usepackage{iccv}
\usepackage{times}
\usepackage{epsfig}
\usepackage{graphicx}
\usepackage{amsmath}
\usepackage{amssymb}

\usepackage[accsupp]{axessibility} 
\usepackage{multirow} 
\usepackage{diagbox}
\usepackage{stackengine}
\usepackage{emptypage}

\usepackage{enumitem}

\usepackage{footmisc}
\usepackage{amsmath}
\usepackage{amsthm}
\usepackage{algorithm}
\usepackage[noend]{algorithmic}

\newtheorem{definition}{Definition}
\newtheorem{proposition}{Proposition}

\usepackage[labelformat=simple]{subcaption}

\usepackage{mathtools}
\DeclarePairedDelimiter{\norm}{\lVert}{\rVert}

\DeclareMathOperator*{\argmin}{arg\,min}
\newcommand\modelName{Limit-Aware Self-Guiding Gradient Sliding Attack\xspace}
\newcommand\modelNameShort{LaS-GSA\xspace}

\usepackage[pagebackref=true,breaklinks=true,letterpaper=true,colorlinks,bookmarks=false]{hyperref}

\iccvfinalcopy % *** Uncomment this line for the final submission

\def\iccvPaperID{5602} % *** Enter the ICCV Paper ID here

\ificcvfinal\fi

\begin{document}

%%%%%%%%% TITLE
\title{Attack as the Best Defense: Nullifying Image-to-image Translation GANs \\ via Limit-aware Adversarial Attack}

\setlist[itemize]{leftmargin=3mm}

\author{Chin-Yuan Yeh\textsuperscript{1,3}, Hsi-Wen Chen\textsuperscript{1}, Hong-Han Shuai\textsuperscript{2}, De-Nian Yang\textsuperscript{3}, Ming-Syan Chen\textsuperscript{1}\\
\textsuperscript{1}National Taiwan University \textsuperscript{2}National Yang Ming Chiao Tung University \textsuperscript{3}Academia Sinica\\
{\tt\small\{d09942009, d09921004, mschen\} @ntu.edu.tw  hhshuai@g2.nctu.edu.tw dnyang@iis.sinica.edu.tw}
}

\maketitle
\ificcvfinal\thispagestyle{empty}\fi

\input{sections/abstract}
\input{sections/Ia_introduction}
\input{sections/II_preliminary}
\input{sections/III_problem_formulation}
\input{sections/IVa_method-limit}
\input{sections/IVb_method-prior}

\input{sections/Va_exp-quant}

\input{sections/Vb_abla-sense}

\input{sections/VI_conclusion}

\section*{Acknowledgmenta}
We thank the National Center for High-performance Computing (NCHC) of National Applied Research Laboratories (NARLabs) in Taiwan for providing computational and storage resources.

{\small
    \bibliographystyle{ieee}
    \bibliography{aref}
}

\cleardoublepage
\appendix
\renewcommand{\theequation}{\thesection.\arabic{equation}}
\setcounter{equation}{0}
\input{appendix/A_proofs/index}

\setcounter{equation}{0}
\setcounter{equation}{0}
\input{appendix/B_more_limit}

\setcounter{equation}{0}
\input{appendix/C_Jacobian}
\setcounter{equation}{0}
\input{appendix/D_distort}
\setcounter{equation}{0}
\input{appendix/E_extended_results}
\end{document}

%% file: sections/abstract.tex
\begin{abstract}
    With the successful creation of high quality image-to-image (Img2Img) translation GANs comes the non-ethical applications of DeepFake and DeepNude. Such misuses of img2img techniques present a challenging problem for society. In this work, we tackle the problem by introducing the \textbf{\modelName (\modelNameShort)}. \modelNameShort follows the \textbf{Nullifying Attack} to cancel the img2img translation process under a black-box setting. In other words, by processing input images with the proposed \modelNameShort before publishing, any targeted img2img GANs can be nullified, preventing the model from maliciously manipulating the images. To improve efficiency, we introduce the \textbf{limit-aware random gradient-free estimation} and the \textbf{gradient sliding mechanism} to estimate the gradient that adheres to the adversarial limit, \ie, the pixel value limitations of the adversarial example. Theoretical justifications validates how the above techniques prevent inefficiency caused by the adversarial limit in both the direction and the step-length. Furthermore, an effective \textbf{self-guiding prior} is extracted solely from the threat model and the target image to efficiently leverage the prior information and guide the gradient estimation process. Extensive experiments demonstrate that \modelNameShort requires fewer queries to nullify the image translation process with higher success rates than $4$ state-of-the-art black-box methods.
\end{abstract}

% Submission system version:
% Due to the great success of image-to-image (Img2Img) translation GANs, many applications with ethics issues arise, e.g., DeepFake and DeepNude, presenting a challenging problem to prevent the misuse of these techniques. In this work, we tackle the problem by a new adversarial attack scheme, namely the Nullifying Attack, which cancels the image translation process and proposes a corresponding framework, the Limit-Aware Self-Guiding Gradient Sliding Attack (LaS-GSA) under a black-box setting. In other words, by processing the image with the proposed LaS-GSA before publishing, any image translation functions can be nullified, which prevents the images from malicious manipulations. First, we introduce the limit-aware RGF and gradient sliding mechanism to estimate the gradient that adheres to the adversarial limit, i.e., the pixel value limitations of the adversarial example. We theoretically prove that our model is able to avoid the error caused by the projection operation in both the direction and the length. Then, an effective self-guiding prior is extracted solely from the threat model and the target image to efficiently leverage the prior information and guide the gradient estimation process. Extensive experiments demonstrate that LaS-GSA requires fewer queries to nullify the image translation process with higher success rates than 4 state-of-the-art methods.

%% file: sections/Ia_introduction.tex
\section{Introduction}
\label{sec:intro}

\begin{figure}[t!]
    \centering
    \includegraphics[width=0.99\linewidth]{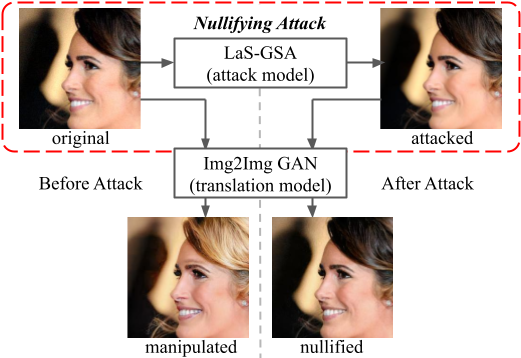}
    \caption{An illustration of nullifying attack against Img2Img GAN. The original portrait is initially manipulated by model \textsc{black2blond} to impaint the portrait image with blond hair.\protect\footnotemark After the nullifying attack, \modelNameShort adds human imperceptible perturbation to the original image and generate an attacked image, which leads the Img2Img GAN to return the nullified image with black hair identical to the original image.}
    \label{fig:architecture}
    \vspace{-2mm}
    \end{figure}

    Recently, Generative Adversarial Networks (GANs)~\cite{goodfellow2014generative} have achieved impressive breakthroughs on various image-to-image translation (Img2Img) tasks, including inpainting~\cite{pathak2016context} and style transfer~\cite{zhu2017unpaired}. These models learn the cross-domain mapping by ensuring that the style of translated images is close to the image style of the target domain while the semantics of the input image are still preserved, \eg, the identity or the layout. 
    \footnotetext{\label{first_footnote}Due to ethical reasons, the illustrative examples utilizes hair-changing model \textsc{black2blond}, instead of DeepNude~\cite{wang2020cnn}. Source code provided in \url{https://github.com/jimmy-academia/LaSGSA}.}
    
    However, Img2Img GANs have also been misused to generate fake images, \ie, DeepFake~\cite{jafar2020forensics} and DeepNude~\cite{lwlodo2019official}. For example, DeepNude excels in undressing full-body shots and producing realistic nude images. Facing the threat of these immoral algorithms, a simple way is to detect DeepFake contents~\cite{rossler2019faceforensics++,wang2020cnn} after the fake images are released. However, even though those post-detection methods can catch the footprints of DeepFake, the manipulated images have already harmed each individual's reputation. Our idea is to defend personal privacy in the first place by nullifying the translation process of misused Img2Img GANs. We aim to attach human-imperceptible perturbations to input images, such that the attacked image can be refrained from being immorally manipulated (to produce obscene images with DeepFake). Thus, our goal is to conduct adversarial attacks against misused Img2Img GANs.
    
    To develop an adversarial attack against misused Img2Img GANs, a simple approach is adopting the \emph{Distorting Attack}~\cite{ruiz2020disrupting,demontis2019adversarial}, which distorts the image translation process of the Img2Img GANs to generate a deteriorated image. However, it can lead to unpredictable results in this case. For example, if the distorting attack is applied to \eg, DeepNude, the distorted regions may appear in the background, and naked images are still created after cropping~\cite{yeh2020disrupting}.
    Therefore, in this paper, we introduce a new attack, namely the \emph{Nullifying Attack}, in a black-box setting.\footnote{Since the white-box attack requires the complete knowledge of the threat model, including the model architectures and weights, we focus on the black-box attack which is more practical in real-world applications (\eg, Google Cloud Vision)~\cite{papernot2017practical,dai2018adversarial}.} Compared with the distorting attack, the nullifying attack is designed to cancel the translation process of misused Img2Img GANs and generate an output image nearly identical to the input one. Figure \ref{fig:architecture} illustrates the nullifying attack, where the targeted Img2Img GAN is nullified by the adversarial example created by our attacked method (detailed later).\footref{first_footnote}

    To facilitate nullifying attack in a black-box setting, one approach is to exploit surrogate models to approximate gradient~\cite{liu2016delving,papernot2017practical,dong2019evading}, \ie, the optimal modification to generate a successful adversarial example. However, preparing surrogate models for an Img2Img GAN requires additional computational resources, and the datasets need to be preprocessed and prepared for model training.\footnote{For instance, training a CycleGAN model involves collecting thousands of relevant images and hundreds of epoch of training on a pair of models with $10^7$ parameters~\cite{zhu2017unpaired}.} Moreover, creating another surrogate model with functions similar to the threat model is morally questionable when the threat models are unethical Img2Img GANs. 
    
    On the other hand, query-based attacks~\cite{chen2017zoo} estimate the gradient for modifying the image by querying the target model and conducting zeroth-order optimization. However, such attacks are inefficient because they usually require more than $10^6$ queries to optimize the adjustment of each pixel for an RGB image. While acceleration schemes have been proposed for the adversarial attack against image classifiers~\cite{tu2019autozoom, andriushchenko2020square}, the adversarial attack against Img2Img GANs is more challenging because it is required to alter the entire output image to a visually distinguishable degree, instead of simply changing a few labels in image classification~\cite{ruiz2020disrupting}.

    To address the above challenge, we introduce \emph{\modelName (\modelNameShort)} to attack Img2Img GANs effectively. First, we prove that naively projecting the gradient, \ie, clipping the gradient~\cite{cheng2019improving, tu2019autozoom} to achieve human-imperceptible modifications, has a detrimental effect on the correctness of the nullified process. Therefore, a \emph{limit-aware strategy} is devised to avoid querying the gradient in the directions that violate the \emph{adversarial limit}, \ie, the pixel value limitations of an adversarial example to follow the imperceptible requirement. Then, a \emph{gradient-sliding mechanism} is introduced to extend the modification along the boundary of the adversarial limit and avoid being trapped in the limit boundary, such that the nullifying attack can be achieved efficiently. Last, by investigating the semantic consistency of Img2Img GANs, we present the \emph{self-guiding prior} that can be extracted from the targeted model directly and remove the cost of preparing surrogate models. At the same time, valuable information is still obtained by the prior to facilitate the nullifying attack in a black-box setting.
    
    The contributions of this paper are as follows:
    \begin{itemize}
        \item We introduce a new adversarial attack on Img2Img GANs, namely the \emph{Nullifying Attack}, and propose the \modelNameShort to cancel the translation process in a black-box setting. 
        \item We investigate the detrimental effects of the projection for the adversarial limit and propose the \emph{limit-aware RGF} and the \emph{gradient sliding mechanism} to effectively mitigate the harm in the gradient estimation process. 
        \item With the \emph{self-guiding prior}, we provide an efficient scheme to extract prior information from Img2Img GANs, removing the need for surrogate models.
        \item Experimental results demonstrate the effectiveness and efficiency of \modelNameShort compared with $4$ state-of-the-art methods on $3$ Img2Img GANs.
    \end{itemize}

%% file: sections/II_preliminary.tex
\section{Preliminary}
\label{sec:preliminary}
\subsection{Image-to-image translation GANs}
    The goal of image-to-image translations~\cite{bashkirova2019adversarial} is to learn a mapping $\mathbb{T}$ that translates an image $x$ from an input domain $X$ to a target domain $Y$, \ie, $\mathbb{T}(x) = y\in Y\,\,\forall x\in X$. As Generative Adversarial Networks (GANs)~\cite{goodfellow2014generative} have been demonstrated to be effective in synthesizing realistic images, Img2Img GANs~\cite{isola2017image,liu2017unsupervised} have been widely adopted to develop state-of-the-art image-to-image translation models. The objective of Img2Img GANs is as follows, 
    \begin{equation}\small
        \min_{G}\max_{D}\mathbb{E}_{x,y}[\log D(x,y)] + \mathbb{E}_{x}[\log(1-D(x, G(x)))],
    \end{equation}
    where the generator $G$ learns to translate $x$ into a realistic target domain sample, and the discriminator $D$ learns to differentiate between a real $y$ and a translated example $G(x)$. While the training is allowed to be conducted either in a supervised setting (\eg, pix2pix~\cite{isola2017image}) or in an unsupervised setting~\cite{liu2017unsupervised}, we first explore the latter due to its higher versatility. CycleGAN~\cite{zhu2017unpaired}, an unsupervised Img2Img GAN, trains a pair of generator $G$ to translate in both directions between the source and target domains. During inference time, we adopt the trained generator $G$ on the specified direction as the targeted translation function $\mathbb{T}$. The nullifying attack is designed to create an adversarial image $x'\approx x_0$ such that $\mathbb{T}$ cannot translate $x'$ to $y\in Y$, but returns the original input $x_0$ nearly unchanged after translation.

\subsection{Projected gradient descent for adversarial attack}
    Given a neural network $f(x)$ and an input-output pair $(x, y)$, the objective of an adversarial attack is to find an adversarial example $x^{\ast}$ that 1) does not generate the expected output $f(x^*) \neq y$, 2) is a legitimate image, and 3) is within the norm-bounded region centering $x$ with a small range $\epsilon \ll 1$ measured in the $\ell_p$ norm,\footnote{$p=2$ or $\infty$ is the common choice for adversarial attacks. In this paper, we adopt $p=\infty$ because it simplifies the projection to pixel-by-pixel numerical upper and lower bounds.} \ie,
    \begin{equation}\small
    \label{eq:opt}
        f(x^*) \neq y, 
        \,\, s.t. \,\, 
       x^*\in[0,1]^N \, \land \,\norm{x^*-x}_\infty \leq \epsilon,
    \end{equation}
    where $N$ is the image dimension, $[0, 1]^N$ is the $N$-orthotope, defined by the legitimate range of values for each pixel (\ie, the \emph{prefix limit}), and $\norm{x^*-x}_\infty \leq \epsilon$ is the $N$-sphere centered at $x$ with radius $\epsilon$ measured in the $\ell_\infty$-norm $\norm{\cdot}$ defined according to the requirement for the perturbation to be human-imperceptible (\ie, the \emph{norm-bound limit}). We denote the union of the two limits as the adversarial limit $\Omega \equiv[0,1]^N \, \land \,\norm{x^*-x}_\infty$ (illustrated in Figure \ref{subfig:advlimit}).
    
    The adversarial example is generated by solving the constrained optimization problem
    \begin{equation}\small
    \label{eq:obj}
        x^*=\argmin_{x'\in\Omega} \mathcal{L}(x'),
    \end{equation}
    where $\mathcal{L}$ is the adversarial loss representing the attack objective, \eg, nullify the functionality of the Img2Img GAN and keep the input unchanged after translation.

    To solve Eq. (\ref{eq:obj}), many gradient-based methods~\cite{carlini2017towards, madry2017towards,szegedy2014intriguing} have been proposed, among which projected gradient descent (PGD) is proven best relying only on first order information~\cite{madry2017towards}. PGD iteratively conducts gradient descent and projection to advance toward the optimal while remaining within the constrained regions. Specifically, let $x^*_{t}$ and $g_t$ denote the adversarial example and the gradient at the $t^\text{th}$ iteration, respectively. The adversarial example at the $t+1^\text{th}$ iteration becomes
    \begin{equation}\small
    \label{eq:pgd}
        x^*_{t+1} = 
        \Pi(x^*_t + \eta g_t), \text{ } g_t = \frac{\nabla\mathcal{L}(x^*_t)}{\norm{\nabla\mathcal{L}(x^*_t)}_2},
    \end{equation}
    where $\Pi$ is the projection operation onto the adversarial limit $\Omega$, \ie, clipping the modification back to the adversarial limit~\cite{papernot2017practical}.
   
\subsection{Black-box setting and random gradient-free estimation}
\label{subsec:rgf}
    Since DeepFake models are generally concealed, nullifying attack naturally occurs in a black-box setting, which only allows one to acquire zeroth-order information, \ie, the system output of a specific query. Therefore, to properly exploit gradient descent optimization, we perform zeroth-order estimations of the gradient by leveraging the Random Gradient-Free (RGF) estimation~\cite{nesterov2017random}. RGF randomly selects query vectors $u_i$ from a unit sphere $\mathcal{U}$ to estimate a gradient $\hat{g_t}$ via
    \begin{equation}\small
        \label{eq:rgf}
        \hat{g}_t  = \frac{1}{q} \sum^q_{i=1}  \frac{L(x^*_t+\delta u_i) - L(x^*_t-\delta u_i)}{2\delta} u_i,  u_i\in\mathcal{U},
    \end{equation}
    where $\delta$ is a small variance. In Eq. (\ref{eq:rgf}), the querying vectors are \emph{flipped} towards the gradient by the multiplication of their own dot product with the gradient. Thus, by querying with radial symmetry, other directions orthogonal to the gradient will be balanced-out in the process to estimate the gradient for nullifying attack effectively.

%% file: sections/III_problem_formulation.tex
\section{Problem formulation}
    For Img2Img GANs, the adversarial attack objective is expressed naturally by shifting the output of the image translation process relative to an attack target $y_\text{target}$, with the corresponding adversarial loss $L_{adv}$ defined as,
    \begin{equation}\small
    \label{eq:loss}
         L_{adv}(x^*) = d(\mathbb{T}(x^*),  y_\text{target}),
    \end{equation}
    where $d$ is the function of squared Euclidean distance, \ie,  $d(x, y) = (\norm{x-y}_2)^2$. By minimizing the loss, the attack model is able to generate an adversarial example $x^*$ that causes the translation function to returns output similar to the target image $y_\text{target}$. In the following, we formally introduce the nullifying attack.\footnote{Compared with attacking a classifier, which only alternates a single output label~\cite{tu2019autozoom}, it is more challenging to attack Img2Img GAN because the attack model is required to ensure the correctness of $10^6$ pixels~\cite{zhu2017unpaired}.} 
    \begin{definition}
    \label{def:null}
        \textbf{Nullifying attack.} The nullifying attack aims to nullify the image translation process such that the adversarial example $x^*$ is mapped back to the original input $x_0$, according to the nullifying loss $L_\text{Null}= \norm{\mathbb{T}(x^*) - x_0}_2^2$.
    \end{definition}
   
    A successful nullifying attack can be adopted as a watermark on personal images such that unethical Img2Img GANs (\eg, DeepNude) cannot manipulate the image.\footnote{We discuss another attack scheme, \emph{Distorting Attack}, which forces the model generates the deteriorated output image in Appendix \ref{appx:distort}.}

%% file: sections/IVa_method-limit.tex
% \section{Limit-aware self-guiding gradient sliding attack}
\section{The \modelNameShort method}
\label{sec:method}
    \input{thefigs/limit}
    In the following, we introduce the \emph{\modelName (\modelNameShort)} scheme, a new black-box adversarial attack, to efficiently nullify the translation process of Img2Img GANs. First, the detrimental effects caused by the projection are investigated, leading to the introduction of the \emph{limit-aware RGF} and the \emph{gradient sliding mechanism}, designed to alleviate the harmful effects. Then, we propose the \emph{self-guiding prior} to fully exploit the threat model for prior information by deriving the approximate solution of the true gradient, removing the requirement for surrogate models or extra datasets~\cite{cheng2019improving}. Last, we present the attack procedure of the \modelNameShort method.
    
\subsection{Limit-aware RGF}

    While the combination of RGF estimation and PGD optimization had been studied in previous black-box attack methods~\cite{tu2019autozoom,brunner2019guessing,cheng2019improving}, they do not consider the detrimental effects of the projecting \ie, clipping, the modification back to the adversarial limit. While the projection is necessary for keeping the adversarial example valid and indistinguishable from the original image, it not only deteriorates the efficiency of the gradient estimation process but also shortens the desired modification towards the estimated gradient, because the projection \emph{pulls back} the out-of-bound gradient. Therefore, the adversarial example is modified towards an undesirable direction, which reduces the effectiveness of both the RGF estimation process and the gradient descent process in PGD. Therefore, we characterize the detrimental effects of projection in twofold: i) misdirection of the gradient, and ii) shortening of the optimization steps.
    
    First, we prove that the projection would mislead the direction of the estimated gradient, harming the efficiency of the nullifying attack process.
    \begin{proposition}
    \label{prop:projection}
        (Proof in Appendix \ref{proof:projection}.) The projection has a detrimental effect on the gradient estimation, \ie, $g\cdot (\Pi(\hat{g}) - \hat{g}) \leq 0$.
    \end{proposition}
    
    To alleviate the detrimental effects of projection, we introduce the \emph{limit-aware RGF} to query the vectors following the adversarial limit, \ie,
    \begin{equation}\small
    \label{eqn:inlimit}
    \Pi(u_i) = u_i \,\forall u_i \rightarrow \Pi(\hat{g}) = \hat{g}.
    \end{equation}
    By examining the convexity of the adversarial limit (detailed in Appendix \ref{proof:limit}), the estimated gradient will not exceed the limit. Based on the observation, we \emph{adjust} the unit $N$-sphere $\mathcal{U}$ in Eq. (\ref{eq:rgf}) to follow the adversarial limit by scaling the basis of $\mathcal{U}$ into a hyperellipsoid $\mathcal{P}$.\footnote{Recall that in Section \ref{subsec:rgf}, the query space of query vector $\mathcal{U}$ is required to exhibit radio symmetry.}

    Concretely, since adversarial limit $\Omega$ forms an $N$-orthotope~\cite{bashkirova2019adversarial,papernot2017practical}, we carefully transform the coordinate system to: 1) set the origin to the current adversarial example $x^*_t$, and 2) adopt every pixel as an independent basis to build an orthonormal basis of the $\mathbb{R}^N$ space. Thus, $\Omega$ becomes an axis-aligned hyperrectangle that includes the origin. Let $\Omega_i$ denote the corresponding range on the $i^{th}$ axis of the $N$-orthotope $\Omega$. To maintain radial symmetry, we define the scale vector $b$ as a vector with the $i^{th}$ element $b_i$ indicating the adjustment range (to increase and decrease the pixel value) for $i^{th}$ pixel, \ie,
    \begin{equation}\small
    \label{eq:limit}
        b_i = (\Omega_i^+, \Omega_i^{-})/2, \,\, \Omega_i^+ \equiv \max(\Omega_i), \,\,\Omega_i^{-} \equiv - \min(\Omega_i),
    \end{equation}
    \begin{equation}\small
    \label{eq:hyperellipsoid}
        \mathcal{P} = \{x \in \mathbb{R}^N : \sum_{i=1}^N \frac{x_i^2}{b_i^2} = 1\}.
    \end{equation}
    As illustrated in Figure \ref{subfig:lmaware}, we scale the unit $N-$sphere into the hyperellipsoid $\mathcal{P}$. Equipped with $\mathcal{P}$, the estimated gradient $\hat{g_t}$ can be formally written as follows,
    \begin{equation}\small
    \label{eq:limitrgf}
        \hat{g_t}  = \frac{1}{q} \sum^q_{i=1} \frac{L_{adv}(x^*_t+\delta u_i) - L_{adv}(x^*_t-\delta u_i)}{2\delta}u_i, \,\,  u_i\in\mathcal{P}.
    \end{equation}
    By the convexity of the adversarial limit, $\hat{g_t}$ satisfies the adversarial limit and maintains radial symmetry by adjusting the range for increasing and decreasing the pixel value simultaneously. Moreover, by adding the scale vector, restricted pixels are effectively squeezed, and thus more adjustments can be facilitated for less restricted pixels.
\subsection{Gradient sliding mechanism}
\label{subsec:gsa}
    In addition to the direction of the estimated gradient, we prove that the projection also shortens the gradient step.
    \begin{proposition}
    \label{prop:length}
        (Proof in Appendix \ref{proof:length}.) The absolute length of the projection result is smaller than the original estimated gradient vector, \ie, $\norm{\Pi(\hat{g})}_2 \leq \norm{\hat{g}}_2$.
    \end{proposition}
    Thus, we propose the \emph{gradient sliding mechanism} to expand each projected gradient step into a series of sliding-steps $\{s_i\}_{i=1}^M$, where $M$ is the number of steps. As illustrated in Figure \ref{subfig:gslide}, instead of being trapped by the adversarial limit, the sliding-steps circumvent along the limit boundary.\footnote{While the gradient step $\Pi(\eta\hat{g}_t)$ is compressed from the estimated gradient $\eta\hat{g}_t$ in Eq. (\ref{eq:rgf}), the sliding-steps $s_i$ (Eq. (\ref{eqn:ministep})) expand the gradient step along the boundary to recover the full length of $\eta\hat{g}_t$.} We carefully configure the steps such that the total length of these sliding-steps is approximately the original gradient step length before projection $l\equiv \norm{\eta\hat{g}}_2$. At step $t+1$, the gradient sliding mechanism starts from the previous adversarial example $x^*_{t}$ and the new adversarial example $\Pi(x^*_t+\eta\hat{g})$ and iteratively derive the next sliding-steps from the previous two sliding-steps, \ie,
    \begin{align}\small
    \label{eqn:ministep}
        &s_1= x^*_t,\,\,s_2= \Pi(x^*_t+\eta\hat{g}),\nonumber\\
        &l_i=\max(0, l-\sum_{k=1}^i\norm{s_k-s_{k-1}}_2),\\
        &s_i = \Pi(s_{i-1} + l_i\cdot(s_{i-1}- s_{i-2})). \nonumber
    \end{align}
    Note that we still adopt projection on the sliding-steps to follow the adversarial limits (detailed in Appendix \ref{proof:mini-on-wall}). The sliding process terminates when the sum of trajectory length exceeds $l$. Since the sliding-step doesn't invoke new queries to the threat model, adopting the gradient sliding mechanism for the nullifying attack does not require additional queries compared with the conventional PGD~\cite{madry2017towards}.

%% file: thefigs/limit.tex
\begin{figure*}
    \centering
    \begin{subfigure}[t]{.42\linewidth}
        \centering
        \includegraphics[width=0.9\linewidth]{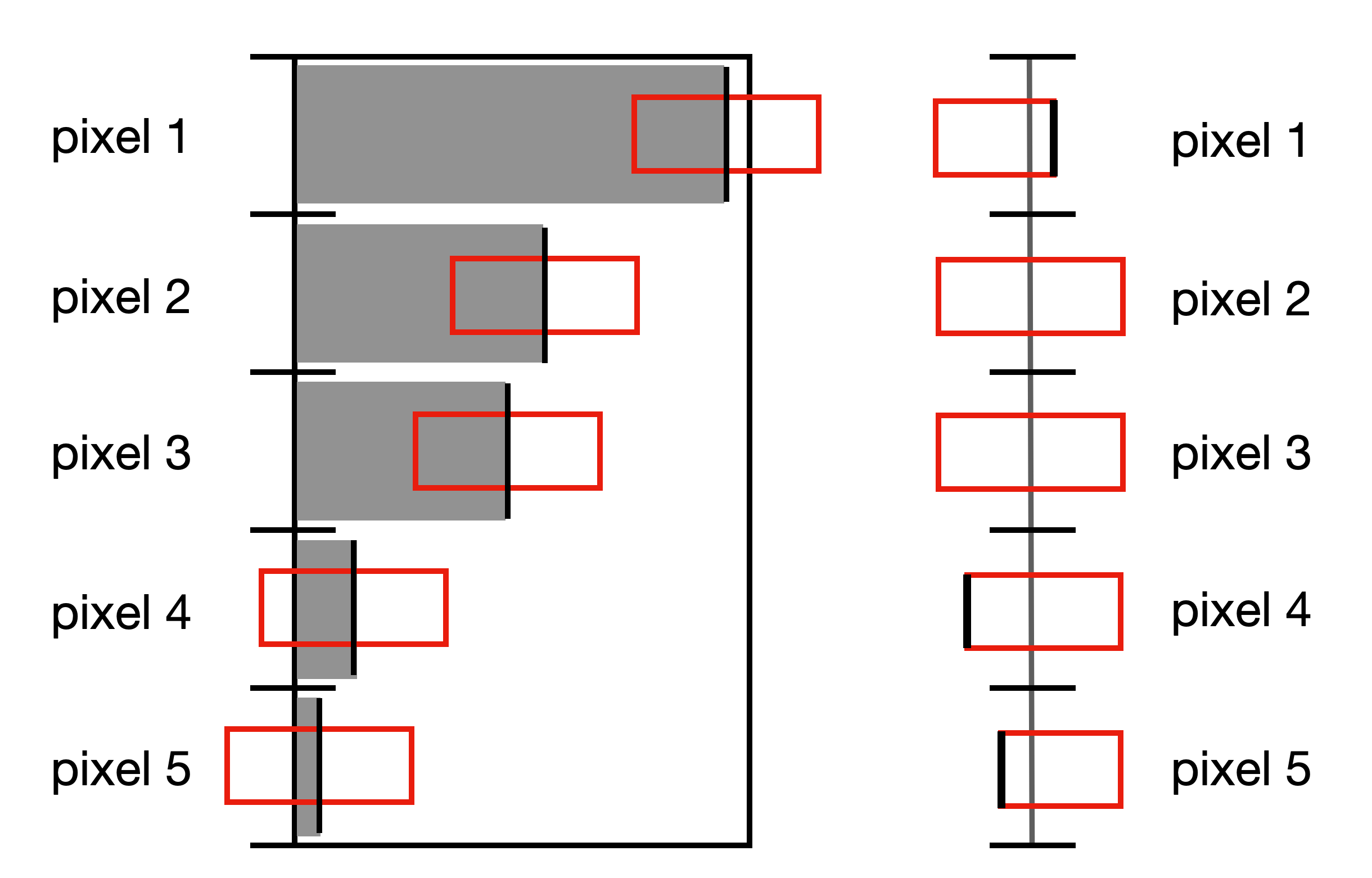}
        \caption{Left: grey bars denote pixel values, the outer black box denotes the \emph{prefix limit} (\eg, $[0,1]$), and the red box denotes the 
        \emph{norm-bound limit} (in $\ell_\infty$). Right: the combined \emph{adversarial limit} is depicted with centered pixel values.}
        \label{subfig:advlimit}
    \end{subfigure}
    \hfill
    \begin{subfigure}[t]{.29\linewidth}
        \centering
        \includegraphics[width=0.9\linewidth]{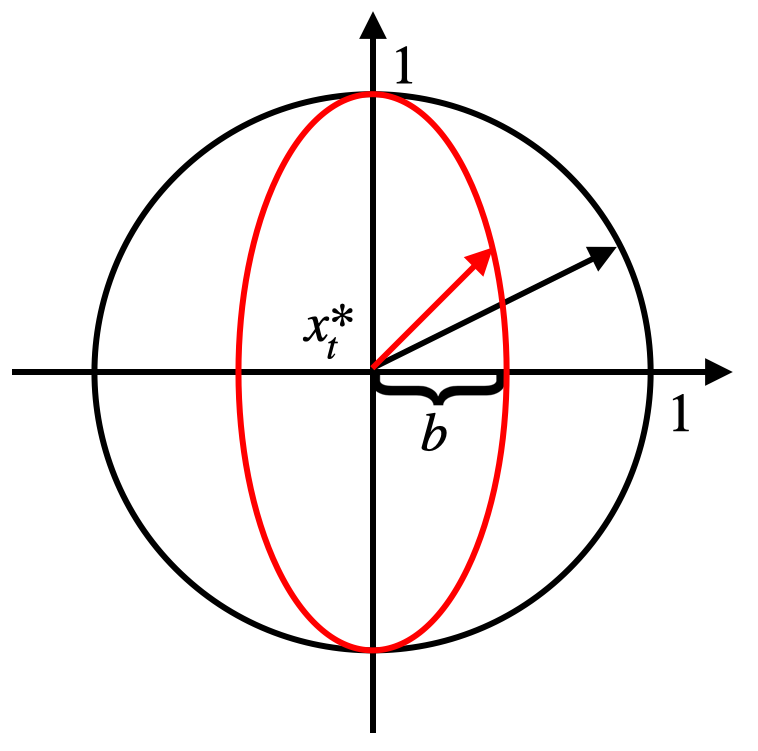}
        \caption{In RGF, random vectors are queried from the unit circle (black). In limit-aware RGF, the queried vector is shifted to the origin-centered ellipse (red).}
        \label{subfig:lmaware}
    \end{subfigure}
    \hfill
    \begin{subfigure}[t]{.23\linewidth}
        \centering
        \includegraphics[width=\linewidth]{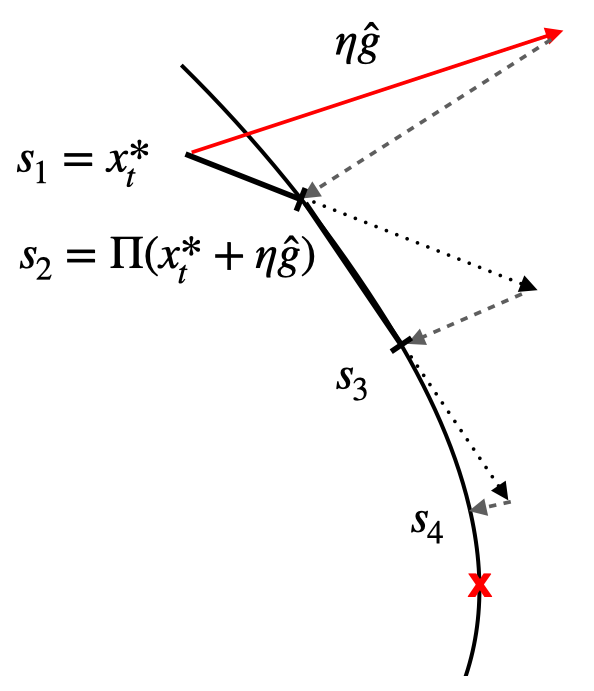}
        \caption{Starting from $s_1 = x^*_t$, PGD moves to $s_2$, whereas gradient sliding selects $s_4$, closer to the optimum (red x).}
        \label{subfig:gslide}
    \end{subfigure}
    \caption{Illustrations of \subref{subfig:advlimit} adversarial limit, \subref{subfig:lmaware} limit-aware RGF, and \subref{subfig:gslide} gradient sliding mechanism.}
    \vspace{-2mm}
\end{figure*}

%% file: sections/IVb_method-prior.tex
\subsection{Self-guiding prior}
\label{subsec:jacobian-prior}
    Although we have addressed the adversarial limit by querying from the limit-aware hyperellipsoid as well as performing the gradient sliding mechanism, nullifying attack is still difficult to be achieved without effective prior information due to the larger search space, \ie, every possible modification of each pixel on the entire image. While several studies~\cite{cheng2019improving,maheswaranathan2018guided} utilize a transfer-based prior that requires a surrogate model trained on extra datasets, it is computationally expensive to prepare a surrogate model. In contrast, by carefully investigating the nullifying process, Img2Img GANs can be exploited as a self-guide because of the semantic consistency of the translation process~\cite{zhu2017unpaired}. 
    
    From Definition \ref{def:null}, the gradient of the nullifying attack at the $t^{th}$ step can be derived as,
    \begin{align}\small\label{eq:gradnull}
        \nabla  L_{Null}(x^*_t) = 2\mathbf{J}^T(\mathbb{T}(x^*_t)-x_0),
    \end{align}
    where the Jacobian matrix transposed $\mathbf{J}^T$ is multiplied to a \emph{discrepancy vector}, \ie, the difference between the current output $\mathbb{T}(x^*_t)$ and the input image $x_0$, the desired change for the adversarial output~\cite{buss2004introduction} (detailed in Appendix \ref{proof:gradient}). 
    
    Due to the semantic consistency of Img2Img GANs~\cite{zhu2017unpaired}, perturbations to each input pixel mostly affect the same pixel in the output~\cite{bashkirova2019adversarial}. Thus, the Jacobian matrix $\mathbf{J}$ is sufficiently diagonal and it is promising to approximate $\mathbf{J}^T$ with $\mathbf{J}$.\footnote{The diagonality of the Jacobian matrix $\mathbf{J}$ is evaluated in Appendix \ref{app:prior}.} Let $a$ denote the discrepancy vector $\mathbb{T}(x^*_t)-x_0$. We estimate the gradient by right multiplying the discrepancy vector to the Jacobian matrix  $\mathbf{J}a$. However, $\mathbf{J}a$ is simply the result of feeding the discrepancy vector into the Img2Img GANs (detailed in Appendix \ref{proof:prior}). We thus arrived at a suitable self-guiding prior $v$,
    \begin{equation}\small
    \label{eq:prior}
    v \equiv \mathbf{J}a 
    \approx \frac{\norm{a}_2 (\mathbb{T}(x^*_t+\delta\hat{a}) - \mathbb{T}(x^*_t))}
    {\delta} , \,\, \hat{a} = \frac{a}{\norm{a}_2}.
    \end{equation}
    With the above approximation, we significantly reduce the time complexity from the $O(N^2)$ to $O(1)$ for the Jacobian transposed $\mathbf{J}^T$ to find a self-guiding prior without exploiting additional surrogate models~\cite{cheng2019improving,maheswaranathan2018guided}, effectively boosting the nullifying attack process.
\input{appendix/algorithm}
%   , which correlates to the nullifying vector direction because $x$ is constrained by the norm-bounded of the original image $x_0$ 
\subsection{Optimization strategy}
\label{subsec:apply-limit-aware}

   Equipped with the limit-aware RGF, the gradient sliding mechanism and the self-guiding prior, we present the final optimization strategy of \emph{\modelName (\modelNameShort)}. Our self-guiding prior is integrated into the RGF and the PGD framework by querying random vectors $u_i$ biased towards the self-guiding prior $v$,
    \begin{equation}\small
    \label{eq:sample}
        u_i = \sqrt{\lambda} \hat{v} + \sqrt{1-\lambda}\hat{t_i}, \,\,
        t_i = \left(\xi_i-(\hat{v}\cdot \xi_i)\hat{v}\right), \xi_i \in\mathcal{P},
    \end{equation} 
    where $\hat{t_i} = \frac{t_i}{\norm{t_i}_2}$, $\hat{v}\equiv\Pi(\frac{v}{\norm{v}})$ is the projected prior, and $\lambda \in [0,1]$ controls the bias of the query $u_i$ towards the prior $\hat{v}$.\footnote{The optimal $\lambda$ is explained and derived in Appendix \ref{app:lambda}.} Each query vector $u_i$ is plugged into Eq. (\ref{eq:rgf}) to estimate the gradient and conduct the PGD process in Eq. (\ref{eq:pgd}). After each gradient step, we perform the sliding-step in Eq. (\ref{eqn:ministep}). Utilizing the three techniques, \modelNameShort effectively and efficiently nullifies the targeted Img2Img GAN model. The pseudocode is presented in Algorithm \ref{alg:lasrgf}.

%% file: appendix/algorithm.tex
\begin{algorithm}[t]
    \caption{\modelName (\modelNameShort)}
    \begin{algorithmic}[1]
    \REQUIRE 
    The translation model $\mathbb{T}$, input image $x_0$, projection operation $\Pi$, sampling variance $\delta$, query number $q$, iteration number $e$, sliding-step number $M$, the learning rate $\eta$.
    \ENSURE The adversarial example $x^*$
    \STATE $x^*\gets x_0$
    \FOR{$i=1$ to $e$}
        \STATE $\hat{a} \gets \Pi(\frac{a}{\norm{a}})$, $a = \mathbb{T}(x^*) - x_0, \hat{g} \gets 0$, 
        \STATE $\hat{v} \gets \Pi(\frac{v}{\norm{v}}), v = \frac{1}{\delta}\left(\mathbb{T}(x^*+\delta\cdot \hat{a}) - \mathbb{T}(x^*)\right)$
        \STATE Find $b$ according to Eq. (\ref{eq:limit})
        \STATE Estimate $\lambda^*$ with $\mathbb{T}$, $\hat{v}$, $q$ according to~\cite{cheng2019improving}
        \FOR{$j=1$ to $q$}
            \STATE Uniform sample $r_j$ from the unit $N$-sphere $\mathcal{U}$;
            \STATE $\xi_j = b\circ r_j$
            \STATE $t_j \gets \xi_j - (\hat{v}\cdot\xi_j)\hat{v}$
            \STATE $u_j = \sqrt{\lambda^*}\hat{v}+\sqrt{1-\lambda^*}t_j$
            \STATE $\hat{g} \gets \hat{g} + \frac{1}{\delta}\left((\mathbb{T}(x^*+\delta u_j) - x_0)^2 - a^2\right)$
        \ENDFOR
        \STATE $x_\text{prev}\gets x^*$, $x_\text{curr} \gets \Pi(x^* +\eta\cdot\frac{1}{q}\hat{g})$,
        \STATE $l\gets\norm{\eta\cdot\frac{1}{q}\hat{g}}_2, \,l_\text{slide} \gets 0$ 
        \FOR{$k=1$ to $M$}
            \STATE $\xi\gets\max(0, l-l_\text{slide})$
            \IF{$\xi=0$}
                \STATE\textbf{break}
            \ENDIF
            \STATE $x_\text{next}\gets \Pi(x_\text{curr}+\xi\cdot(x_\text{prev}-x_\text{curr}))$
            \STATE $x_\text{prev} \gets x_\text{curr}$
            \STATE $x_\text{curr} \gets x_\text{next}$
            \STATE $l_\text{slide} \gets l_\text{slide} + \norm{x_\text{curr} - x_\text{prev}}_2$
        \ENDFOR
        \STATE $x^* \gets x_\text{curr}$
    \ENDFOR
    \RETURN $x^*$
    \end{algorithmic}
    \label{alg:lasrgf}
\end{algorithm}

%% file: sections/Va_exp-quant.tex
\section{Experiments}
\label{sec:experiment}
    We compare \modelNameShort with $4$ state-of-the-art black-box adversarial attack schemes. All attack methods are implemented for $3$ Img2Img GANs relevant to the manipulation of personal images: $2$ trained on closed-up portraits and $1$ trained on full-body shots. We first present the experiment setup. Then, we present quantitative and qualitative evaluations of the attack results and an ablation study. 

\subsection{Experimental setup}
\label{subsec:implementation-detail}
\paragraph{Threat Models.}
    We adopt CycleGAN~\cite{liu2017unsupervised} as the default Img2Img GAN architecture for the following threat models: 1) \textsc{black2blond}, which is trained on HQ-CelebA dataset~\cite{karras2017progressive} to translate people with black hair to blonde hair, 2) \textsc{none2glasses}, which adds glasses to portraits, also trained on HQ-CelebA dataset, and 3) \textsc{blue2red}, which is trained on self-prepared datasets of clean images for people wearing blue and red shirts from Google Image Search for translating blue shirts to red shirts. Besides, we select $100$ testing samples~\cite{taori2019targeted} that are i.i.d. to the training set of each threat model.\footnote{Additional qualitative results of nullifying attack on the $3$ Img2Img GANs, \ie, \textsc{black2blond}, \textsc{none2glasses} and \textsc{blue2red} and distorting attack on $3$ models, \ie, \textsc{str2seg}, \textsc{facade2label}, and \textsc{night2day}, are presented in Appendix \ref{app:extensive}.}
\vspace{-2mm}
\paragraph{Baselines.}
    The proposed \textit{\modelNameShort} is compared with $4$ state-of-the-art methods. 1) \textit{Bandit}~\cite{ilyas2018prior} adopts the time-dependent prior vector to guide the sampling process. 2) \textit{Square}~\cite{andriushchenko2020square} performs localized square-shaped updates at random positions. 3) \textit{RGF}~\cite{nesterov2017random} randomly samples the query vectors from the unit $N$-sphere. 4) \textit{Prior-RGF}~\cite{cheng2019improving} utilize the surrogate model to bias the query vectors in \textit{RGF} towards the transfer-based prior vector estimated from the surrogate model.\footnote{The surrogate models are trained with the same architecture and procedure on $100$ i.i.d. samples of the original training set.}
    The querying variance $\delta$, norm-bound $\epsilon$, and learning rate $\eta$ are set to $0.001$, $0.1$, and $1$, respectively. To provide transfer priors for the Prior-RGF method, surrogate models are prepared for each threat model with the \emph{same} architectures and conditions.
\vspace{-2mm}
\paragraph{Evaluations.}
    To evaluate the results of different attack schemes, we present a task-oriented score, \ie, the nullifying score $s_\text{Null}$,
    \begin{align}\small
    \label{eqn:score}
        s_{\text{Null}}(x^*) = \left[1- \frac{(\norm{\mathbb{T}(x^*)-x_0}_2)^2}{\norm{y_0 - x_0}_2^2}\right]\times 100,
    \end{align}
    where the original translation distance $\norm{y_0 - x_0}_2, y_0 = \mathbb{T}(x_0)$ acts as a normalization. Following~\cite{yeh2020disrupting}, we consider adversarial examples $x^*$ successful if $ s_{\text{Null}}(x^*)$ is greater than the threshold $75$.\footnote{The threshold is determined by $100$ samples with $50$ users~\cite{yeh2020disrupting}.} The attack success rate (ASR) is defined as the percentage of the successful attack on test images in $100,000$ query budgets. The query count (Q) represents the average number of attempted queries (stopping upon passing the threshold) for each example.
\input{sections/figures}
\subsection{Quantitative evaluations}
    \label{subsec:quantitative}
   Table \ref{tab:full-compare} compares the proposed \modelNameShort against baseline methods in terms of the attack success rate (ASR) and the query count (Q) of the $100$ testing images for each threat model. For all threat models, \modelNameShort outperforms all the other approaches in both ASR and Q. Remarkably, Bandit attack could not pass the threshold score within a $100,000$ query budget for some threat models. Compared to RGF, \modelNameShort also consistently achieves better performance in both ASR and Q by at least $10\%$, because \modelNameShort carefully examines the clipping effect and exploits self-guiding prior to attack the CycleGAN effectively. Even though Prior-RGF is equipped with a surrogate model, which has the identical CycleGAN structure trained on an i.i.d. testing dataset to estimate the prior, \modelNameShort still outperforms Prior-RGF by $2\%$ to $17\%$ regarding ASR. This is because the output space of GANs is much larger than image classifiers~\cite{yeh2020disrupting}, and thus transferring the gradient across different models is much more challenging.
   
\subsection{Ablation Study}
    Table \ref{tab:abla-compare} presents the ablation studies on \textsc{black2blond} with RGF and four variants of our method, including 1) \textit{GSA}: with only the gradient sliding mechanism, 2) \textit{S-RGF}: with only the self-guiding prior, 3) \textit{S-GSA}: with both the self-guiding prior and the gradient sliding mechanism, and 4) \textit{LaS-RGF}: with both the limit-aware RGF and the self-guiding prior. First, variants equipped with the gradient sliding mechanism (*-GSA) consistently improve the performance by at least $3\%$ regarding ASR compared with RGF. Besides, the self-guiding prior increases the overall ASR to $80\%$, and the limit-aware RGF improves ASR to $85\%$. Furthermore, the results in the query count (Q) follow a similar trend, in which \modelNameShort reduces by $20\%$ of queries compared to RGF. Notice that \modelNameShort outperforms LaS-RGF by $17.2\%$ regarding query efficiency since the gradient sliding mechanism carefully estimates the adversarial limit and prolongs the optimization steps along the constraint boundary, leading to better efficiency.

%% file: sections/figures.tex
\begin{table*}
\begin{tabular}{cc}
\begin{minipage}[h]{.68\linewidth}
\centering
    \begin{tabular}{|c|cc|cc|cc|}
    \hline
        \multirow{2}{*}{\backslashbox{Methods}{Models}}
        & \multicolumn{2}{c|}{\textsc{black2blond}}
        & \multicolumn{2}{c|}{\textsc{none2glasses}}
        & \multicolumn{2}{c|}{\textsc{blue2red}} \\
        \cline{2-7}
        & ASR & Q/(s) & ASR & Q/(s) & ASR & Q/(s)  \\
    \hline
        Bandit~\cite{ilyas2018prior} &0\%&(2)   & 10\%& 90,019 &0\%&(5)\\
        Square~\cite{andriushchenko2020square} &23\%&87,196 & 40\% &60,194  & 25\%&80,778\\
        RGF~\cite{nesterov2017random} &71\%&53,237 &88\%&58,049  &23\%&85,969 \\
        Prior-RGF~\cite{cheng2019improving} &69\%&51,330 &78\%&81,580  &23\%& 80,463\\
        \hline
        \modelNameShort &\textbf{85\%}&\textbf{42,917} &\textbf{95\%}&\textbf{40,298} &\textbf{40\%}&\textbf{79,934}\\
    \hline
    \end{tabular}
    \caption{Quantitative results of the black-box attack against Img2Img GANs with a limit of 100,000 queries. We report the attack success rate (ASR) and the query count (Q) for all $100$ test samples. If the attack fails in all $100$ test samples, the average final score (s) is presented with parentheses.}
    \label{tab:full-compare}
\end{minipage}&
\begin{minipage}{.29\linewidth}
\begin{tabular}{|c|cc|}
    \hline
        Methods & ASR & Q \\
    \hline
        RGF &71\%&53,237\\
        \hline
        GSA &76\%&48,849\\
        S-RGF &81\%&49,743\\
        S-GSA  &84\%& 43,694\\
        LaS-RGF  &80\%&51,871\\
        \hline
        \modelNameShort &\textbf{85\%}&\textbf{42,917} \\
    \hline
    \end{tabular}
    \caption{Ablation test results for \textsc{black2blond} with the attack success rate (ASR) and the query count (Q).}
    \label{tab:abla-compare}
\end{minipage}
\end{tabular}
\end{table*}

\begin{figure*}[t]
    \centering
    	\begin{subfigure}{.135\linewidth}
            \centering
            \includegraphics[width=0.99\linewidth]{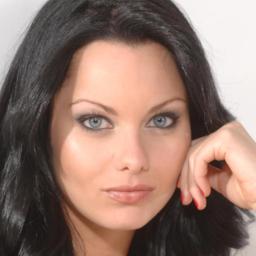}
            \includegraphics[width=0.99\linewidth]{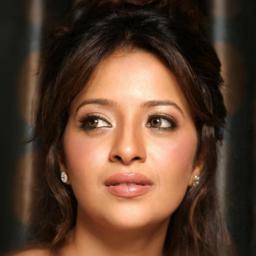}
            % \stackunder[5pt]{\includegraphics[width=0.99\linewidth]{cvimg/compare/blond/14-ax0.jpg}}{input}
            \stackunder[5pt]{\includegraphics[width=0.99\linewidth]{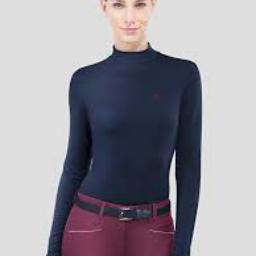}}{input}
    	\end{subfigure}
    	    \begin{subfigure}{.135\linewidth}
            \centering
            \includegraphics[width=0.99\linewidth]{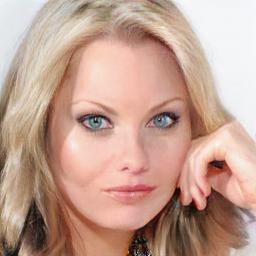}
            \includegraphics[width=0.99\linewidth]{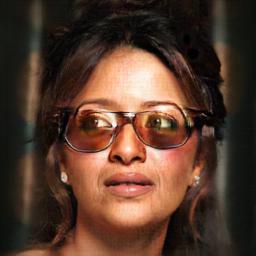}
    % 		\stackunder[5pt]{\includegraphics[width=0.99\linewidth]{cvimg/compare/blond/14-by0.jpg}}{expected}
    		\stackunder[5pt]{\includegraphics[width=0.99\linewidth]{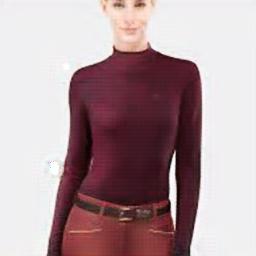}}{expected}
    	\end{subfigure}
        \begin{subfigure}{.135\linewidth}
            \centering
            \includegraphics[width=0.99\linewidth]{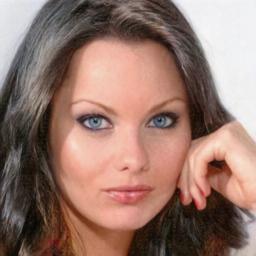}
            \includegraphics[width=0.99\linewidth]{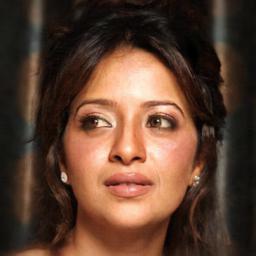}
    % 		\stackunder[5pt]{\includegraphics[width=0.99\linewidth]{cvimg/compare/blond/14-mine.jpg}}{\modelNameShort}
    		\stackunder[5pt]{\includegraphics[width=0.99\linewidth]{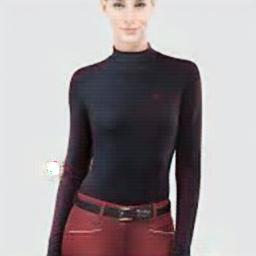}}{\modelNameShort}
    	\end{subfigure}
	    \begin{subfigure}{.135\linewidth}
            \centering
            \includegraphics[width=0.99\linewidth]{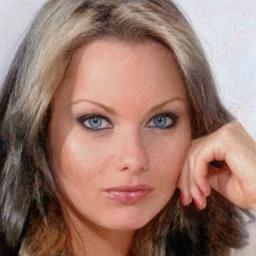}
            \includegraphics[width=0.99\linewidth]{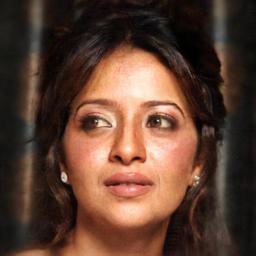}
            % \stackunder[5pt]{\includegraphics[width=0.99\linewidth]{cvimg/compare/blond/14-prgf.jpg}}{Prior-RGF}
            \stackunder[5pt]{\includegraphics[width=0.99\linewidth]{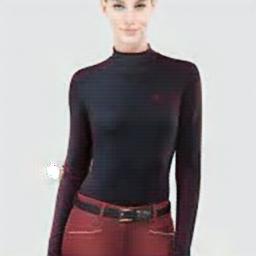}}{Prior-RGF}
    	\end{subfigure}
        \begin{subfigure}{.135\linewidth}
            \centering 
            \includegraphics[width=0.99\linewidth]{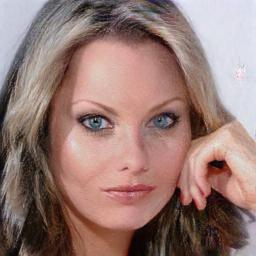}
            \includegraphics[width=0.99\linewidth]{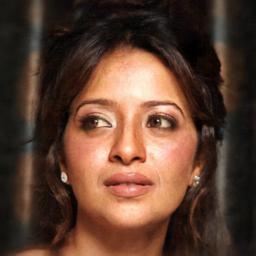}
            % \stackunder[5pt]{\includegraphics[width=0.99\linewidth]{cvimg/compare/blond/14-rgf.jpg}}{RGF}
            \stackunder[5pt]{\includegraphics[width=0.99\linewidth]{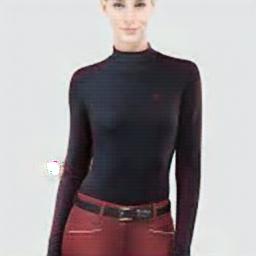}}{RGF}
    	\end{subfigure}
	    \begin{subfigure}{.135\linewidth}
            \centering
            \includegraphics[width=0.99\linewidth]{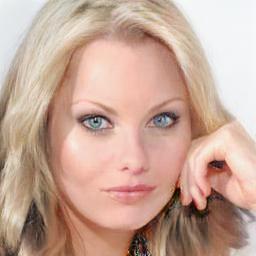}
            \includegraphics[width=0.99\linewidth]{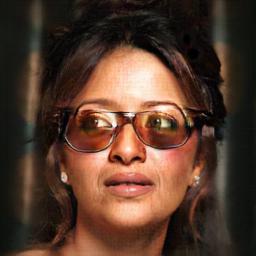}
            % \stackunder[5pt]{\includegraphics[width=0.99\linewidth]{cvimg/compare/blond/bandbl.jpg}}{Bandit}
            \stackunder[5pt]{\includegraphics[width=0.99\linewidth]{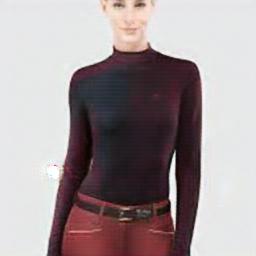}}{Bandit}
    	\end{subfigure}
        \begin{subfigure}{.135\linewidth}
            \centering
            \includegraphics[width=0.99\linewidth]{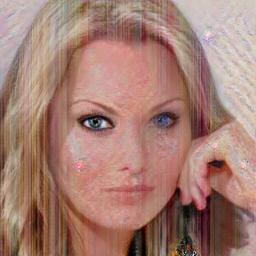}
            \includegraphics[width=0.99\linewidth]{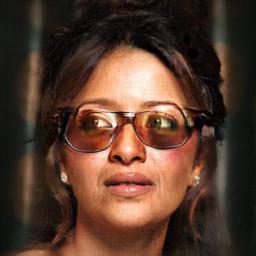}
            % \stackunder[5pt]{\includegraphics[width=0.99\linewidth]{cvimg/compare/blond/blsq.jpg}}{Square}
            \stackunder[5pt]{\includegraphics[width=0.99\linewidth]{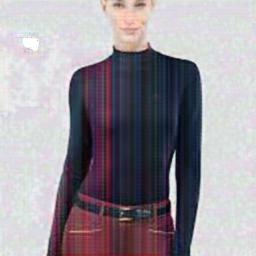}}{Square}
    	\end{subfigure}	
        \caption{Comparing attack methods with adversarial results for model \textsc{black2blond}.}
        \label{fig:qual-compare}
\end{figure*}

\begin{figure*}[!t]\small
    \centering
    \begin{subfigure}{.132\linewidth}
    	\stackunder[5pt]{\includegraphics[width=.98\linewidth]{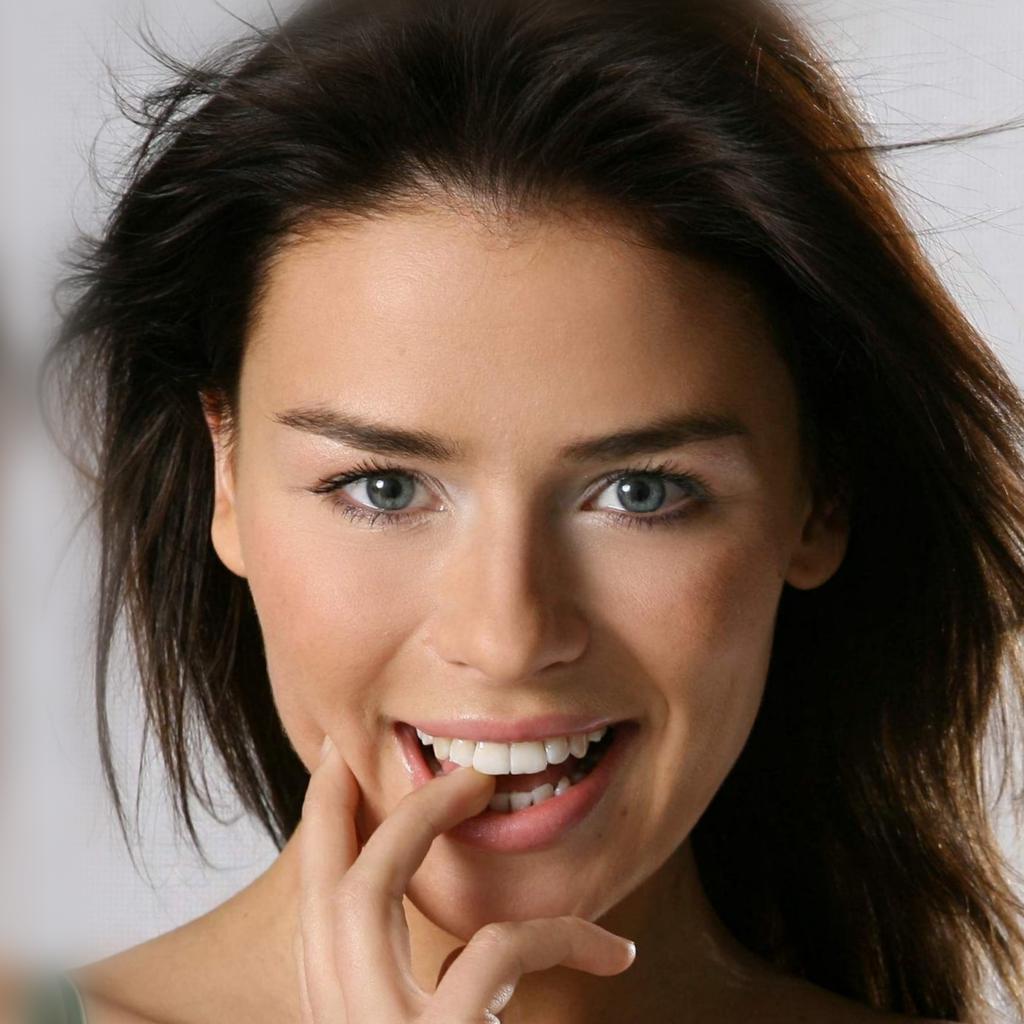}}{input}
    \end{subfigure}
    \begin{subfigure}{.132\linewidth}
    	\stackunder[5pt]{\includegraphics[width=.98\linewidth]{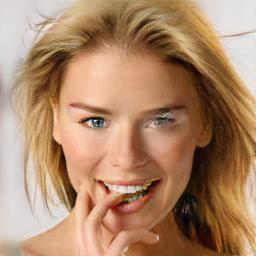}}{expected}
    \end{subfigure}
    \begin{subfigure}{.132\linewidth}
        \centering
		\includegraphics[width=.98\linewidth]{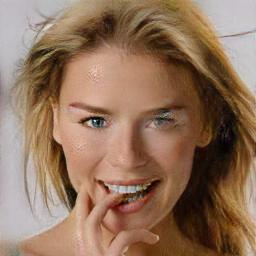}
        {$40/1,056$}
	\end{subfigure}
	\begin{subfigure}{.132\linewidth}
        \centering
        \includegraphics[width=.98\linewidth]{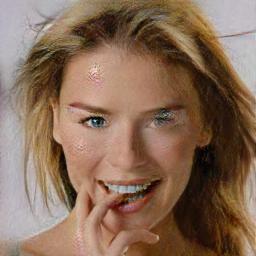}
        {$54/6,331$}
	\end{subfigure}
	\begin{subfigure}{.132\linewidth}
        \centering
        \includegraphics[width=.98\linewidth]{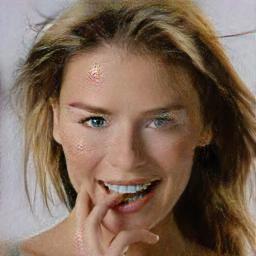}
        {$67/10,551$}
	\end{subfigure}
	\begin{subfigure}{.132\linewidth}
        \centering
        \includegraphics[width=.98\linewidth]{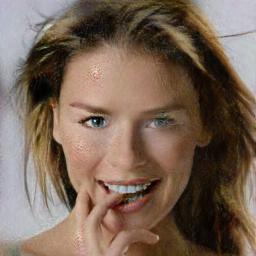}
        {$75/31,651$}
	\end{subfigure}
	\begin{subfigure}{.132\linewidth}
        \centering
	    \includegraphics[width=.98\linewidth]{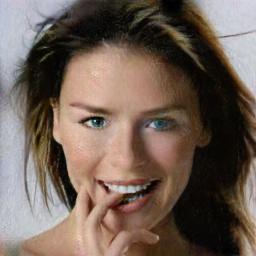}
        {$94/100,000$}
	\end{subfigure}
	
% 	\begin{subfigure}{.132\linewidth}
%     	\stackunder[5pt]{\includegraphics[width=.98\linewidth]{cvimg/blue/7-x0.jpg}}{input}
%     \end{subfigure}
%     \begin{subfigure}{.132\linewidth}
%     	\stackunder[5pt]{\includegraphics[width=.98\linewidth]{cvimg/blue/7-y0.jpg}}{expected}
%     \end{subfigure}
%     \begin{subfigure}{.132\linewidth}
%         \centering
% 		\includegraphics[width=.98\linewidth]{cvimg/blue/7-y-15-3103.jpg}
%         {$15/3,103$}
% 	\end{subfigure}
% 	\begin{subfigure}{.132\linewidth}
%         \centering
%         \includegraphics[width=.98\linewidth]{cvimg/blue/7-y-35-11375.jpg}
%         {$35/11,375$}
% 	\end{subfigure}
% 	\begin{subfigure}{.132\linewidth}
%         \centering
% 	    \includegraphics[width=.98\linewidth]{cvimg/blue/7-y-55-23783.jpg}
%         {$55/23,783$}
% 	\end{subfigure}
% 	\begin{subfigure}{.132\linewidth}
%         \centering
%         \includegraphics[width=.98\linewidth]{cvimg/blue/7-y-75-40327.jpg}
%         {$75/40,327$}
% 	\end{subfigure}
% 	\begin{subfigure}{.132\linewidth}
%         \centering
%         \includegraphics[width=.98\linewidth]{cvimg/blue/7-finaly.jpg}
%         {$90/99,401$}
% 	\end{subfigure}
    \caption{The attack process of \modelNameShort on \textsc{black2blond}. From left to right: the original input image, expected Img2Img GAN output, and intermediate results of \modelNameShort attack, with the score ($s_\text{Null}$)/query number shown below.}
    \label{fig:process}
\end{figure*}

\begin{figure*}[t]
    \centering
    	\begin{subfigure}{.153\linewidth}
            \centering
    		\includegraphics[width=0.99\linewidth]{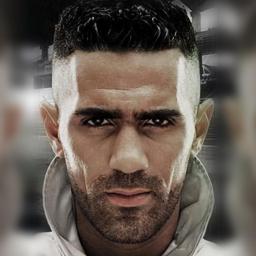}
    		\includegraphics[width=0.99\linewidth]{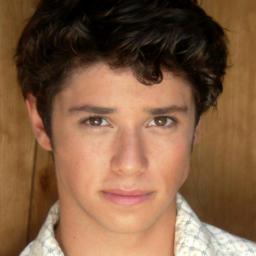}
    		\stackunder[5pt]{\includegraphics[width=0.99\linewidth]{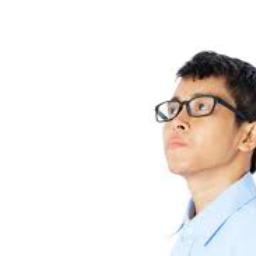}}{input}
    % 		\stackunder[5pt]{\includegraphics[width=0.99\linewidth]{cvimg/blue/0-x0.jpg}}{input}
    	\end{subfigure}
    	    \begin{subfigure}{.153\linewidth}
            \centering
    		\includegraphics[width=0.99\linewidth]{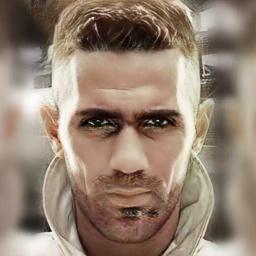}
    		\includegraphics[width=0.99\linewidth]{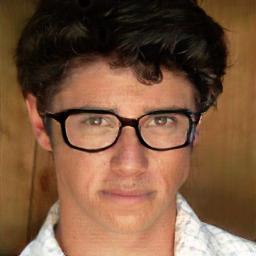}
    		\stackunder[5pt]{\includegraphics[width=0.99\linewidth]{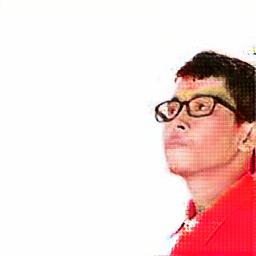}}{expected}
    % 		\stackunder[5pt]{\includegraphics[width=0.99\linewidth]{cvimg/blue/0-y0.jpg}}{expected}
    	\end{subfigure}
    % 	\begin{subfigure}{.113\linewidth}
    %         \centering
    % % 		\includegraphics[width=0.99\linewidth]{cvimg/general/str2seg/0-x0.jpg}
    % % 		\includegraphics[width=0.99\linewidth]{cvimg/general/str2seg/17-x0.jpg}
    % 		\includegraphics[width=0.99\linewidth]{cvimg/general/blond/2-x0.jpg}
    % 		\includegraphics[width=0.99\linewidth]{cvimg/general/blond/77-x0.jpg}
    % 		\includegraphics[width=0.99\linewidth]{cvimg/glasses/17-finalx.jpg}
    % 		\includegraphics[width=0.99\linewidth]{cvimg/glasses/15-finalx.jpg}
    % 		\includegraphics[width=0.99\linewidth]{cvimg/general/redblue/11-finalx.jpg}
    % % 		\stackunder[5pt]{\includegraphics[width=0.99\linewidth]{cvimg/general/redblue/11-finalx.jpg}}{adversarial}
    % 		\stackunder[5pt]{\includegraphics[width=0.99\linewidth]{cvimg/general/redblue/10-finalx.jpg}}{adversarial}
    % 	\end{subfigure}
        \begin{subfigure}{.153\linewidth}
            \centering
    		\includegraphics[width=0.99\linewidth]{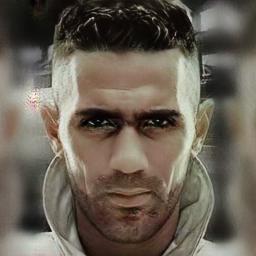}
    		\includegraphics[width=0.99\linewidth]{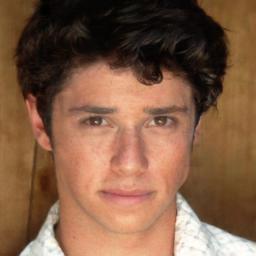}
    		\stackunder[5pt]{\includegraphics[width=0.99\linewidth]{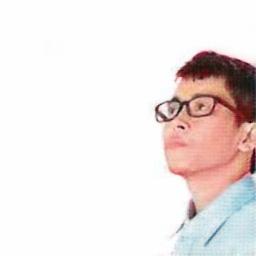}}{\modelNameShort}
    % 		\stackunder[5pt]{\includegraphics[width=0.99\linewidth]{cvimg/blue/0-finaly.jpg}}{\modelNameShort}
    	\end{subfigure}
    	\begin{subfigure}{.153\linewidth}
            \centering
    		\includegraphics[width=0.99\linewidth]{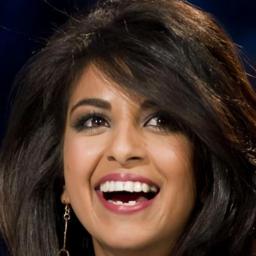}
    		\includegraphics[width=0.99\linewidth]{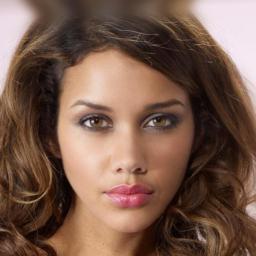}
    		\stackunder[5pt]{\includegraphics[width=0.99\linewidth]{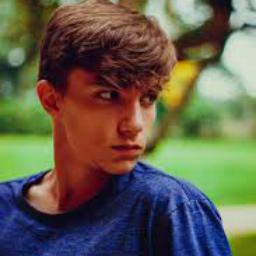}}{input}
    % 		\stackunder[5pt]{\includegraphics[width=0.99\linewidth]{cvimg/blue/1-x0.jpg}}{input}
    	\end{subfigure}
    	    \begin{subfigure}{.153\linewidth}
            \centering
    		\includegraphics[width=0.99\linewidth]{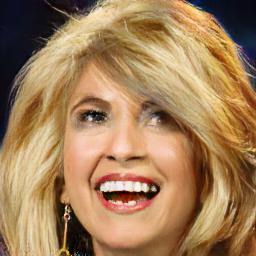}
    		\includegraphics[width=0.99\linewidth]{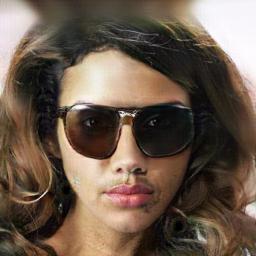}
    		\stackunder[5pt]{\includegraphics[width=0.99\linewidth]{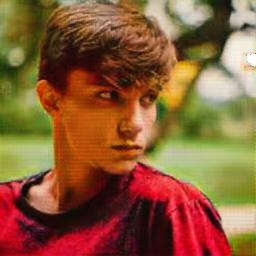}}{expected}
    % 		\stackunder[5pt]{\includegraphics[width=0.99\linewidth]{cvimg/blue/1-y0.jpg}}{expected}
    	\end{subfigure}
    % 	\begin{subfigure}{.113\linewidth}
    %         \centering
    %         % \includegraphics[width=0.99\linewidth]{cvimg/general/str2seg/3-ax0.jpg}
    % % 		\includegraphics[width=0.99\linewidth]{cvimg/general/str2seg/5-x0.jpg}
    % 		\includegraphics[width=0.99\linewidth]{cvimg/general/blond/70-x0.jpg}
    % 		\includegraphics[width=0.99\linewidth]{cvimg/general/blond/36-ax0.jpg}
    % 		\includegraphics[width=0.99\linewidth]{cvimg/glasses/23-finalx.jpg}
    % 		\includegraphics[width=0.99\linewidth]{cvimg/glasses/16-finalx.jpg}
    % 		\includegraphics[width=0.99\linewidth]{cvimg/general/redblue/12-finalx.jpg}
    % % 		\stackunder[5pt]{\includegraphics[width=0.99\linewidth]{cvimg/general/redblue/12-finalx.jpg}}{adversarial}
    % 		\stackunder[5pt]{\includegraphics[width=0.99\linewidth]{cvimg/general/redblue/2-finalx.jpg}}{adversarial}
    % 	\end{subfigure}
        \begin{subfigure}{.153\linewidth}
            \centering
    		\includegraphics[width=0.99\linewidth]{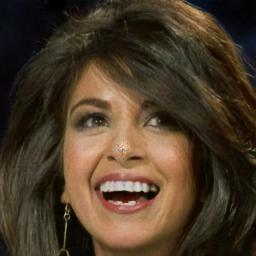}
    		\includegraphics[width=0.99\linewidth]{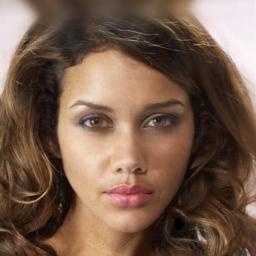}
    		\stackunder[5pt]{\includegraphics[width=0.99\linewidth]{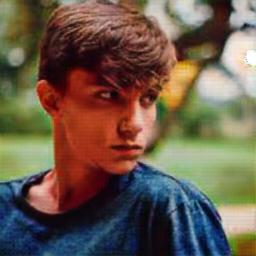}}{\modelNameShort}
    % 		\stackunder[5pt]{\includegraphics[width=0.99\linewidth]{cvimg/blue/1-finaly.jpg}}{\modelNameShort}
    	\end{subfigure}
        \caption{Qualitative \modelNameShort attack results against model \textsc{black2blond},  \textsc{none2glasses}, and \textsc{red2blue} from top to bottom, each presenting the input images, expected output of each model, and the final output after applying \modelNameShort.}
        \label{fig:full-example}
\end{figure*}

% ($1^\text{st}-2^\text{nd}$ row),
% ($3^\text{rd}-4^\text{th}$ row) and

%% file: sections/Vb_abla-sense.tex
\subsection{Qualitative evaluation}
\label{subsec:qualitative}
    Figure \ref{fig:qual-compare} compares the visual quality of the attack results. Consistent with the quantitative results, Bandit and Square fail to alter the image output. Square blurs the image with the vertical stripes because their mechanism favors rectangle perturbations. For \textsc{black2blond}, while other methods only slightly modify the hair color to \textit{brown}, \modelNameShort is the only one that nullifies the translation process and returns a black hair image because it effectively utilizes the limit-aware gradient estimation and the gradient sliding mechanism to ensure the correctness of nullifying process in both direction and length. While \modelNameShort achieves similar results on \textsc{none2glasses} and \textsc{blue2red} compared with RGF and Prior-RGF, as shown in Table \ref{tab:full-compare}, it requires fewer queries because the self-guiding prior can provide meaningful guide for the modification direction. 
    % The result demonstrates that \modelNameShort can protect personal pictures against malicious applications (\eg, DeepNude) effectively.
    Figure \ref{fig:process} visualizes the nullifying process of \modelNameShort on \textsc{black2blond}, which recovers the hair color from blond to black. As query counts increase, the resulting output image shifts from blond hair back to black. We observe that a nullifying score $s_\text{Null}=75$ is sufficient, with up to $30,000$ queries. Nonetheless, with $100,000$ queries, the nullifying attack can make the adversarial output much closer to the original input. 
    Finally, Figure \ref{fig:full-example} further presents $2$ additional samples for each threat model to demonstrate the generality of \modelNameShort. Our limit-aware strategy follows the adversarial limit and keep the adversarial perturbations imperceptible. With the nullifying attack scheme, \modelNameShort effectively creates adversarial examples that cause models \textsc{black2blond}, \textsc{none2glasses}, and \textsc{blue2red} to generate output images that are almost identical to the original input images, canceling the respective functionality. More qualitative results are presented in Appendix~\ref{app:extensive}. 

%% file: sections/VI_conclusion.tex
\section{Conclusion}
\label{sec:conclusion-and-discussion}
    In this work, we introduce a new adversarial attack on Img2Img GANs in a black-box setting, namely \emph{Nullifying Attack}, to defend against malicious applications (\eg, DeepFake). We propose the \emph{\modelName (\modelNameShort)} method, which incorporates the \emph{limit-aware RGF}, the \emph{gradient sliding mechanism}, and the \emph{self-guiding prior} to cancel the image translation process of Img2Img GANs. Experimental results demonstrate the effectiveness and efficiency of our proposed method in $3$ different translation processes. Future work includes reducing the vulnerability of Img2Img GANs against adversarial attacks for safety-critical applications.

%% file: appendix/A_proofs/index.tex
\section{Proofs}
\label{app:proof}
\input{appendix/A_proofs/i_proposition1}

\input{appendix/A_proofs/ii_proposition2}
\input{appendix/A_proofs/iv_prior}

%% file: appendix/A_proofs/i_proposition1.tex
\subsection{Proof of Proposition \ref{prop:projection}}
\label{proof:projection}

\paragraph{Proposition \ref{prop:projection}}\textit{The projection has a detrimental effect on the gradient estimation, i.e., $g\cdot (\Pi(\hat{g}) - \hat{g}) \leq 0$.}

    \begin{proof}
    We first prove that the inner product between the true gradient $g$ and the RGF estimated gradient $\hat{g}$ is non-detrimental, \ie, $g\cdot\hat{g}\geq 0$. According to  Eq. (\ref{eq:rgf}), the estimated gradient is  $\hat{g}_t = \frac{1}{q} \sum^q_{i=1}\frac{\partial  L(x^*_t)}{\partial u_i} u_i,$ where $u_i$ is drawn from a radial symmetrical distribution $\mathcal{U}$, \eg, the unit $N$-sphere. The partial derivative of $L(\cdot)$ at point $x$, with regard to $u$ can be formally written as
        \begin{equation}\small\label{eq:partial}
            \frac{\partial  L(x)}{\partial u}
            = \frac{d}{d\zeta} L(x+\zeta u)|_{\zeta=0},
        \end{equation}
    where $\zeta$ is a dummy variable. Eq. (\ref{eq:partial}) can be further expressed by the multivariable chain rule,
    \begin{align}\small\label{eq:dummy}
        \frac{d}{d\zeta} L(x+\zeta u)
        &= \sum_j\frac{\partial L(x+\zeta u)}{\partial (x+\zeta u)_j}
        \frac{\partial (x+\zeta u)_j}{\partial \zeta} \nonumber\\
        &= \sum_j \frac{\partial L(x+\zeta u)}{\partial e_j}u_j,
    \end{align}
    where $e_j$ is the unit vector along the $j^\text{th}$ basis. Since the dummy variable $\zeta$ is set to $0$, according to Eq. (\ref{eq:dummy}), we find \begin{align}\small
    \frac{\partial  L(x)}{\partial u} 
    &= \sum_j \frac{\partial  L(x)}{\partial e_j}u_j 
    = \left(\sum_j \frac{\partial  L(x)}{\partial e_j}e_j\right) \cdot u \nonumber\\
    &\equiv \nabla  L(x) \cdot u.
    \end{align}
    Therefore, $g \cdot \hat{g}_t$ can be rewritten as,
    \begin{equation}\small\label{eq:nodetrimental}
    \begin{split}
        g \cdot \hat{g}_t 
        &= \nabla L(x)\cdot 
        \left(
        \sum^q_{i=1}\frac{\partial  L(x)}{\partial u_i} u_i 
        \right)\\
        &= \nabla L(x)\cdot 
        \left(\sum^q_{i=1} (\nabla L(x)\cdot u_i)u_i\right)\\
        &= \sum^q_{i=1} (\nabla L(x)\cdot u_i)^2 
        \geq 0.
    \end{split}
    \end{equation}

    Next, we prove $\hat{g}\cdot (\Pi(\hat{g}) - \hat{g}) \leq 0$ by decomposing the projection function $\Pi$ into the \emph{prefix limit} projection function $\Pi_{\Omega}$ and the \emph{norm-bound limit} projection function $\Pi_{\mathcal{B}(x, \epsilon)}$, \ie,  
    \begin{equation}\small
        \Pi(g) = \Pi_{\mathcal{B}(x, \epsilon)}\circ\Pi_{\Omega}(g).
    \end{equation}
    
    For the prefix limit, since $x^*$ is the optimal solution according to Eq. (\ref{eq:opt}), $\norm{x^*-x}\leq\epsilon$ holds. Therefore, if $\Pi_{\Omega}(\hat{g}) \neq \hat{g}$, we have
    \begin{equation}\small\label{eq:inequal}
        \norm{x^*-x+\hat{g}} \geq \epsilon \geq \norm{x^*-x+\Pi_{\Omega}(\hat{g})},
        \end{equation}
    and $\hat{g}\cdot(\Pi_{\Omega}(\hat{g}) - \hat{g})$ becomes 
        \begin{equation}\small
        \begin{split}
        &(x^*-x+\hat{g})\cdot((x^*-x+\Pi_{\Omega}(\hat{g})) - (x^*-x+\hat{g})) \\
        &- (x^* - x) \cdot(\Pi(\hat{g}) - \hat{g}).
        \label{eq:pro1}
        \end{split}
        \end{equation}
The first term in Eq. (\ref{eq:pro1}) is smaller or equal to zero by the following simplification.
        \begin{equation}\small
        \begin{split}
            & (x^*-x+\hat{g})\cdot((x^*-x+\Pi_{\Omega}(\hat{g})) - (x^*-x+\hat{g})) \\
            &=( x^*-x+\hat{g})\cdot(x^*-x+\Pi_{\Omega}(\hat{g})) - (x^*-x+\hat{g})^2 \\
            & \leq  \norm{x^*-x+\hat{g}}\norm{x^*-x+\Pi_{\Omega}(\hat{g})} - \norm{x^*-x+\hat{g}}^2 \\
            & \leq 0.
        \end{split}
    \end{equation}
    Besides, the absolute value of the second term, \ie, $(x^* - x) \cdot(\Pi(\hat{g}) - \hat{g})$, is smaller than the first term because $\norm{x^*-x}\leq\epsilon\leq\norm{x^*-x+\hat{g}}$, and thus $\hat{g} \cdot (\Pi_{\Omega}(\hat{g}) - g) \leq 0$. Note that if $\Pi_{\Omega}(\hat{g}) = \hat{g}$, then $\hat{g}\cdot (\Pi_{\Omega}(\hat{g}) - \hat{g})=0$. Therefore, $\hat{g} \cdot (\Pi_{\Omega}(\hat{g}) - \hat{g}) \leq 0$.
    
    Afterwards, we prove that the norm-bound limit projection has a detrimental effect on the gradient estimation. Without loss of generality, we assume that $p=2$.\footnote{If $p=\infty$, the proof is the same as that of the prefix limit.} Each pixel of the adversarial example $x^*_i$ has an upper limit $b^u_i$ and a lower limit $b^l_i$, representing the valid range of adjustment, \ie, $(x_i^*)_{t+1} - x_i \in [b_i^l,b_i^u]$.\footnote{It is worth noting that $b^l_i \leq 0 \leq b^u_i$, otherwise $(x^*_i)_t$ does not exist.} Then, $\Pi_{\mathcal{B}(x, \epsilon)}(\hat{g})_i$ can be expressed as $\min(\max(\hat{g}_i, b^l_i), b^u_i)$, \ie,

    \begin{equation}\small
    \begin{split}
        \Pi_{\mathcal{B}(x, \epsilon)}(\hat{g})_i &= \min(\max(\hat{g}_i, b^l_i), b^u_i)
        \\&=\begin{cases}
            b^u_i, & \hat{g}_i > b^l_i \,\,\text{and}\,\, \hat{g}_i > b^u_i\\
            \hat{g}_i, & \hat{g}_i \geq b^l_i \,\,\text{and}\,\, \hat{g}_i \leq b^u_i \\
            b^l_i, & \hat{g}_i < b^l_i
        \end{cases}
        \\&=
        \begin{cases}
        b^u_i, & \hat{g}_i > b^u_i\\
        \hat{g}_i, & b^l_i  \leq \hat{g}_i \leq b^u_i \\
        b^l_i, & \hat{g}_i < b^l_i
        \end{cases}.
    \end{split}
    \end{equation}
    Therefore, $\hat{g}\cdot (\Pi(\hat{g}) - \hat{g})$ can be written as,
        \begin{equation}\small
        \begin{split}
            \hat{g}\cdot (\Pi_{\mathcal{B}(x, \epsilon)}(\hat{g}) - \hat{g}) &= \sum_i \hat{g}_i \cdot (\Pi_{\mathcal{B}(x, \epsilon)}(\hat{g}) - \hat{g})_i 
            \\&= 
            \begin{cases}
                \hat{g}_i\cdot (b^u_i - \hat{g}_i), & \hat{g}_i > b^u_i \geq 0 \\
                0, & b^l_i  \leq \hat{g}_i \leq b^u_i \\
                \hat{g}_i \cdot (b^l_i - \hat{g}_i), & \hat{g}_i < b^l_i \leq 0
            \end{cases}.
        \end{split}
        \end{equation}
    Since each element of  $\hat{g}\cdot (\Pi_{\mathcal{B}(x, \epsilon)}(\hat{g}) - \hat{g})$ is non-positive, $\hat{g}\cdot (\Pi(\hat{g}) - \hat{g}) \leq 0$. Moreover, as two projection operations $\Pi_{\Omega}$ and $\Pi_{\mathcal{B}(x, \epsilon)}$ are monotonic, the projection function $\Pi$ that combines the two operations still satisfies the property, \ie, $\hat{g}\cdot (\Pi(\hat{g}) - \hat{g}) \leq 0$. Finally, since $g\cdot\hat{g}\geq 0$, the inequality becomes
    \begin{equation}
        g\cdot (\Pi(\hat{g}) - \hat{g}) \leq (g\cdot \hat{g}) (\hat{g}  \cdot (\Pi(\hat{g}) - \hat{g})) \leq 0.
    \end{equation}
    The proposition follows.
\end{proof}

%% file: appendix/A_proofs/ii_proposition2.tex
\subsection{Proof of Proposition \ref{prop:length}}
    \label{proof:length}
    \paragraph{Proposition \ref{prop:length}}\textit{The length of the projection operation result is smaller than the original estimated gradient vector, i.e., $\norm{\Pi(\hat{g})}_2 \leq \norm{\hat{g}}_2$.}
    \begin{proof}
        Recall that under $\ell_\infty$ norm, the adversarial limit can be regarded as an $N$-orthotope containing the origin, with the limit in each basis defined as Eq. (\ref{eq:limit}). Intuitively, gradient vector will reduce in length after being \emph{trimmed off} at the limit. For $\ell_2$ norm, the adversarial limit is the union of a $\ell_2$ $N$-sphere and an $N$-orthotope. If the projected gradient is on the $\ell_2$ $N$-sphere, the gradient's length (prior to projection) is larger or equal to the radius of the $N$-sphere, and thus $\norm{\Pi(\hat{g})}_2 \leq \norm{\hat{g}}_2$.
    \end{proof}

\subsection{Proof of Eq. (\ref{eqn:inlimit})}
    \label{proof:limit}
    \paragraph{Eq. (\ref{eqn:inlimit})}
    $\Pi(u_i) = u_i \,\forall u_i \rightarrow \Pi(\hat{g}) = \hat{g}.$
    \begin{proof}
        Recall that the estimated gradient is the weighted sum of the query vectors. The space within the adversarial limit is convex since it is the union of convex spaces ($N$-orthotope $\Omega$ and hypersphere $\mathcal{B}_{p}(x, \epsilon)$). Thus, by convexity the estimated gradient using query vectors within the adversarial limit will remain in the adversarial limit. 
    \end{proof}
 
\subsection{Proof of ending statement in Section \ref{subsec:gsa}}
\label{proof:mini-on-wall}
    \paragraph{Description}\textit{The gradient sliding process follows the adversarial limit boundary.}
    \begin{proof}
    
    According to Eq. (\ref{eqn:ministep}), starting from $s_2$, sliding steps are the results of projection operations, and we have $s_i = \Pi(n) = \Pi(\Pi(n)) = \Pi(s_i)$. Therefore, each $s_i$ is within the adversarial limit for $i\geq2$.
    \end{proof}

%% file: appendix/A_proofs/iv_prior.tex
\subsection{Proof of (\ref{eq:gradnull})}
\label{proof:gradient}
    \paragraph{Description of Eq. (\ref{eq:gradnull})}\textit{Given the nullifying loss $L_\text{Null}= - (\norm{\mathbb{T}(x^*) - x}_2)^2$, the gradient is,}
    \begin{align}\small
        &\nabla  L_\text{Null}(x^*_t) = -2\mathbf{J}^T(\mathbb{T}(x^*_t)-x_0).\nonumber
    \end{align}
    
\begin{proof}
    We expand $L_\text{Null}$ for each pixel component and simplify the variables with $x=x^*_t$ and $y=x_0$ as 
    \begin{equation}\small
        L_\text{Null} = -(\norm{\mathbb{T}(x)-y}_2)^2 = -\sum_i^N (\mathbb{T}(x)_i-y_i)^2.
    \end{equation}
    By the definition of the gradient operator,
    \begin{align}\small
        \nabla L_\text{Null}(x) &=-\sum_j^N \frac{\partial}{\partial e_j} \left(\sum_i^N (\mathbb{T}(x)_i-y_i)^2 \right)\hat{e_j}\nonumber\\
        &= -2\sum_{i,j}^N (\mathbb{T}(x)_i-y_i)\frac{\partial\mathbb{T}(x)_i}{\partial e_j}\hat{e_j}.
        \label{eqn:grad-expanded}
    \end{align}
    However, the element $\mathbf{J}_{ij}$ of the Jacobian matrix is $\frac{\partial\mathbb{T}(x)_i}{\partial e_j}$. Therefore, the $j^\text{th}$ element of Eq. (\ref{eqn:grad-expanded}) can be regarded as the result of the element-wise multiplication of the $j^\text{th}$ column vector of $\mathbf{J}$ and the vector $\mathbb{T}(x)-y$. By transposing $\mathbf{J}$, column vectors are transposed into row vectors. Therefore, the multiplication along with $\hat{e_j}$, summed over $j$, becomes the right-multiplication of the column vector $\mathbb{T}(x)-y$ to the transposed Jacobian matrix $\mathbf{J}^T$. Thus,
    \begin{equation}\small
        \sum_{i,j}^N (\mathbb{T}(x)_i-y_i)\frac{\partial\mathbb{T}(x)_i}{\partial e_j}\hat{e_j} \equiv \mathbf{J}^T(\mathbb{T}(x)-y).
    \end{equation}
    Eq. (\ref{eq:gradnull}) is proved.
\end{proof}
    
\subsection{Proof of Eq. (\ref{eq:prior})}
\label{proof:prior}
    \paragraph{Description of Eq. (\ref{eq:prior})}\textit{Given the Img2Img translation model $\mathbb{T}$ and the discrepancy vector $a$, the self-guiding prior vector $v\equiv \mathbf{J}a$ can be approximated as}
    \begin{equation*}\small
        v \equiv \mathbf{J}a 
        \approx \frac{\norm{a}_2(\mathbb{T}(x^*_t+\delta\hat{a}) - \mathbb{T}(x^*_t)}{\delta} , \,\, \hat{a} = \frac{a}{\norm{a}_2}.
    \end{equation*}
    
\begin{proof}
    By the definition of the Jacobian matrix,
    \begin{equation}\small\label{eq:priorapprox}
        \begin{split}
        v & \equiv \mathbf{J}a
            = \frac{\partial\mathbb{T}_j}{\partial e_i} a_i 
            = \frac{d}{d\zeta}\mathbb{T}(x+\zeta e_i)_j|_{\zeta=0}\cdot a_i
            \\
        &  = \lim_{\delta\rightarrow0}
            \frac{1}{\delta} 
            \left(
            \mathbb{T}(x^*_t + (\zeta+\delta) e_i)_j|_{\zeta=0} - \mathbb{T}(x+\zeta e_i)_j|_{\zeta=0}
            \right)a_i,
        \end{split}
    \end{equation}
    where $e_i$ represents the $i^\text{th}$ basis, $\zeta$ is a dummy variable, and $\delta$ is the infinitesimal value for the derivation. We first expand Eq. (\ref{eq:priorapprox}) into a Taylor series, \ie,
    
     \begin{equation}\small\label{eq:taylor}
        \begin{split}
        v  & = \lim_{\delta\rightarrow0}
            \frac{1}{\delta} 
            \Bigg(
            \mathbb{T}(x^*_t + \zeta e_i)_j|_{\zeta=0}
            + \frac{\partial \mathbb{T}(x^*_t + \zeta e_i)_j|_{\zeta=0}}{\partial e_i} \delta e_i 
            \\&+ O(\delta^2)
            - \mathbb{T}(x^*_t + \zeta e_i)_j|_{\zeta=0} \bigg)a_i, \\
        \end{split}
    \end{equation}
    where $O(\delta^2)$ denotes the higher-order terms. Without loss of generality, we approximate $v$ by dropping the higher-order term $O(\delta^2)$. Since $a_i$ is a scalar which can be placed into the numerical limit, we rewrite Eq. (\ref{eq:taylor}) as, 
    \begin{align}\small
        v  \approx  \lim_{\delta\rightarrow0}&
            \frac{1}{\delta} 
            \left(
            \frac{\partial \mathbb{T}(x^*_t + \zeta e_i)_j|_{\zeta=0}}{\partial e_i} \delta e_i
            \right)a_i \nonumber\\
        %  = \lim_{\delta\rightarrow0}&
        %     \frac{1}{\delta} 
        %     \left(
        %     \frac{\partial \mathbb{T}(x^*_t + \zeta e_i)_j|_{\zeta=0}}{\partial e_i}
        %     \delta\frac{a_i e_i}{\norm{a}_2}
        %     \right)\norm{a}_2 \nonumber\\
         = \lim_{\delta\rightarrow0}&
            \frac{1}{\delta} 
            \left(
            \frac{\partial \mathbb{T}(x^*_t + \zeta e_i)_j|_{\zeta=0}}{\partial e_i}
            \delta\hat{a}
            \right)\norm{a}_2 \nonumber\\
          = \lim_{\delta\rightarrow0}&
            \frac{1}{\delta} 
            \left(
            \mathbb{T}(x^*_t+\delta\hat{a})_j
            % \nonumber\\ & 
            - \mathbb{T}(x^*_t)_j
            \right)\norm{a}_2.
    \end{align}
    By replacing the limit with a small $\delta$, the approximation of Eq. (\ref{eq:prior}) holds.
\end{proof}

%% file: appendix/B_more_limit.tex
\section{Implementation detail of \modelNameShort method}
\label{app:lambda}
    In this section, we detail the prior guiding query framework and the selection of the optimal $\lambda$. Then, we extend the framework to support limit aware RGF. The radial symmetry property is also presented.

\subsection{Prior guiding query and the optimal \texorpdfstring{$\lambda$}~}
\label{app:cone-shape-approx}
    
    Following~\cite{cheng2019improving}, the random query vector $u_i$ in Eq. (\ref{eq:rgf}) is biased towards the prior vector $v$ by a hyperparameter $\lambda\in(1,0)$, which can be expressed by
    \begin{equation}\small
        \label{apeq:query-vector-u}
            u_i = \sqrt{\lambda} \hat{v} + \sqrt{1-\lambda}\hat{t_i}, \,\,
            t_i = \left(\rho_i-(\hat{v}\cdot \rho_i)\hat{v}\right),\,\, \rho_i\in\mathcal{U},
    \end{equation} 
    where $\hat{v}$ and $\hat{t}_i$ represent the normalization of the vectors $v$ and $t_i$, respectively. Note that the above equation can be regarded as projecting the random unit vector onto a \emph{cone} revolving the prior vector $v$, since $\rho_i\in\mathcal{U}$. 
    
    The optimal $\lambda$ minimizes the difference between the estimated and the true gradient~\cite{tu2019autozoom}, \ie,
    \begin{equation}\small\label{eq:lambdaloss}
        \lambda^* = \argmin_\lambda \,\, \mathbb{E}[\norm{\nabla L(x)-\hat{g}}_2^2].
    \end{equation}
    
    By the Pythagorean theorem, we rewrite Eq. (\ref{eq:lambdaloss}) by minimizing the vector component of the true gradient \emph{orthogonal} to the expected estimated gradient $\mathbb{E}[\hat{g}])$, 
    \begin{equation}\small
    \label{apeq:minimization}
        \lambda^* = \argmin_\lambda \,\,
        \left(
        \norm{\nabla L(x)}^2 - \left(\frac{\nabla L (x)\cdot\mathbb{E}[\hat{g}]}{\norm{\mathbb{E}[\hat{g}]}_2}\right)^2
        \right),
    \end{equation}
    where the second term is the vector component of the real gradient \emph{parallel} to $\mathbb{E}[\hat{g}]$. As the estimated gradient $\hat{g}$ is the weighted average of the query vector $u_i$, $\hat{g}$ is expressed as,  
    \begin{equation}\small
        \hat{g}=\frac{1}{q}\sum_i(\nabla L(x)\cdot u_i) u_i \equiv \frac{1}{q}\sum_i\hat{g_i}.
    \end{equation}
    \begin{equation}\small
    \label{apeq:expectation-estimate}
        \mathbb{E}[\hat{g}] =\mathbb{E}[\hat{g_i}] = \mathbb{E}[u_i u_i^T]\nabla L(x),
    \end{equation}
    where $u_i^T$ denotes the transpose of the query vector $u_i$. By replacing $u_i$ in Eq. (\ref{apeq:expectation-estimate}) with Eq. (\ref{apeq:query-vector-u}),
    \begin{equation}\small
    \label{apeq:uut-lambda}
        \mathbb{E}[u_i u_i^T] = \mathbb{E}[\lambda\hat{v}\hat{v}^T + (1-\lambda)\hat{t_i}\hat{t_i}^T], \,\,
        t_i = \left(\xi_i-(\hat{v}\cdot \xi_i)\hat{v}\right).
    \end{equation}
    Since $t_i$ is orthogonal to $v$, $\hat{t_i}$ is a unit vector in both the original $\mathbb{R}^N$ vector space and the $\mathbb{R}^{N-1}$ vector space orthogonal to $v$. Following~\cite{cheng2019improving}, we decompose $\hat{t_i}$ as $\sum_{j=1}^{N-1}a_j e_j$, where $e_j$ denotes a vector basis orthogonal to $\hat{v}$. With $\sum e_j e_j^T = \mathbf{I} - \hat{v}\hat{v}^T$ and $\sum{{a_j}^2} = 1$, $\mathbb{E}[\hat{t_i}\hat{t_i}^T]$ becomes
    \begin{equation}\small
    \label{apeq:tt-i-vv}
    \begin{split}
        \mathbb{E}[\hat{t_i}\hat{t_i}^T] &= 
        \mathbb{E}\left[\left(\sum_{j=1}^{N-1}a_j e_j\right)\left(\sum_{k=1}^{N-1}a_k e_k^T\right)\right]
        \\&= \mathbb{E}[{a_j}^2]\sum e_j e_j^T = \frac{1}{N-1}(\mathbf{I} - \hat{v}\hat{v}^T).
    \end{split}
    \end{equation}
    Besides, we also expand $\mathbb{E}[\norm{\hat{g}}_2^2]$ by
    \begin{equation}\small
    \label{apeq:expectation-norm-squared}
    \begin{split}
        \mathbb{E}[\norm{\hat{g}}_2^2] 
        &= \mathbb{E}[\norm{\hat{g} - \mathbb{E}[\hat{g}]}_2^2] + \norm{\mathbb{E}[\hat{g}]}_2^2 
        \\&=\frac{1}{q} \mathbb{E}[\norm{\hat{g_i} - \mathbb{E}[\hat{g_i}]}_2^2] + \norm{\mathbb{E}[\hat{g_i}]}_2^2 
        \\&=\frac{1}{q} \mathbb{E}[\norm{\hat{g_i}}] + (1 - \frac{1}{q})\norm{\mathbb{E}[\hat{g_i}]}_2^2.        
    \end{split}
    \end{equation}
    After inserting Eq. (\ref{apeq:expectation-estimate}) and Eq. (\ref{apeq:expectation-norm-squared}) into Eq. (\ref{apeq:minimization}),
    \begin{equation}\small
    \begin{split}
    \lambda^* &= \argmin_\lambda 
        {\norm{\nabla L(x)}_2}^2\bigg(
            1 - \frac{\text{NUM}}{\text{DENOM} }
        \bigg),\\
        \text{NUM} &= (\lambda\alpha^2+\frac{1-\lambda}{N-1}(1-\alpha^2))^2,\\
        \text{DENOM} &= (1-\frac{1}{q})(\lambda^2\alpha^2+(\frac{1-\lambda}{N-1})^2(1-\alpha^2))\\
        &+\frac{1}{q}(\lambda\alpha^2+\frac{1-\lambda}{N-1}(1-\alpha^2)) ,      
    \end{split}
    \end{equation}
    where $\alpha$ denotes the cosine similarity between the estimated gradient and the true gradient, \ie, $\alpha =  \frac{v\cdot\nabla  L(x)}{\norm{v}_2\norm{\nabla  L(x)}_2}$. To find the maximum of the second term, we set its derivative with regard to $\lambda$ as $0$ and obtain $\lambda^*$ as follows.
    \begin{equation}\small
    \label{apeq:optimal-lambda}
        \lambda^* = 
        \begin{cases}
        0 & \text{if} \,\, \alpha^2\leq\cfrac{1}{N+2q-2} \\
        1 & \text{if} \,\, \alpha^2 \geq \cfrac{2q-1}{N+2q-2}\\
        \cfrac{(1-\alpha^2)(\alpha^2(N+2q-2) - 1)}
            {2\alpha^2Nq-\alpha^4N(N+2q-2) - 1} & 
            \text{otherwise}. 
        \end{cases}
    \end{equation}

    Since the cosine similarity between $u_i$ and $v$ is regulated by $\sqrt{\lambda}$, if $\alpha\approx1$, the estimated gradient $\hat{g}$ is similar to the true gradient and $\lambda\approx1$; otherwise, if $\alpha\approx0$, $\lambda\approx0$, and the problem is reduced to estimating the gradient $\nabla L(x)$ in the $N-1$ dimension excluding the dimension along $v$.

    Finally, since the inner product of a vector and the gradient can be approximated into the directional derivative,
    \begin{equation}\small
    \label{apeq:estimate-alpha}
        \alpha = 
            \frac{v\cdot\nabla  L(x)}{\norm{v}\norm{\nabla L(x)}} 
            = \frac{1}{\norm{\nabla  L(x)}} \frac{\partial L(x)}{\partial\hat{v}}.
    \end{equation}
    On the other hand, the squared cosine similarity of two random vectors is expected to be $\frac{1}{N}$. Thus, by estimating the inner product squared and extracting out the gradient length term, the gradient length $\norm{\nabla L(x)}$ can be calculated as,
    \begin{align}\small
    \label{apeq:estimate-normlength}
        \frac{1}{S}\sum_{i=1}^S\left(\frac{\partial  L(x)}{\partial \rho_i}\right)^2
        &= \norm{\nabla  L(x)}^2\frac{1}{S}\sum_{i=1}^S\left(\frac{\nabla  L(x)}{\norm{\nabla  L(x)}}\cdot\rho_i\right)^2
        \nonumber\\&\approx \frac{ \norm{\nabla  L(x)}^2}{N}, \,\, \rho_i\in\mathcal{U},
    \end{align}
    where $\rho_i$ is queried from the unit $N$-sphere $\mathcal{U}$, and $S$ determines the number of vectors to query. Furthermore, the partial derivative in Eq. (\ref{apeq:estimate-alpha}) can be estimated as
    \begin{equation}\small
        \frac{\partial  L(x)}{\partial\hat{v}} \approx \frac{1}{\delta}( L(x+\delta\hat{v})-  L(x)), \,\, \delta \ll 1.
    \end{equation}
    Thus, we can adopt the cosine similarity $\alpha$ to estimate the optimal $\lambda^*$ in Eq. (\ref{apeq:optimal-lambda}).

\subsection{Limit-aware projection to the prior-guiding query}

    Appendix \ref{app:cone-shape-approx} presents an effective method to exploit the information from a prior vector. Here, we extend the biased querying strategy to support the limit-aware RGF. First, Eq. (\ref{apeq:query-vector-u}) is modified by replacing the prior $\hat{v}$ with the projected prior $\Pi(\hat{v})$ and replacing the query vector $r_i \in \mathcal{U}$ with $\xi_i$ in the hyperellipsoid $\mathcal{P}$. Thus, according to Eq. (\ref{eq:sample}), we still obtain the optimal cone as described in the $N$-sphere case. Besides, $\mathcal{P}$ is scaled symmetrically to fit in the adversarial limit from the unit $N$-sphere $\mathcal{U}$. Thus, the new probability distribution on the cone is still symmetric and satisfies the radial symmetry requirement (detailed in Appendix \ref{app:rgf}).
    
    The new query vectors are given as
    \begin{equation}\small
        u_i = \sqrt{\lambda} \hat{v} + \sqrt{1-\lambda}\hat{t_i}, \,\,
        t_i = \left(\xi_i-(\hat{v}\cdot \xi_i)\hat{v}\right), \xi_i \in\mathcal{P},
    \end{equation} 
     where $\hat{t_i} = \frac{t_i}{\norm{t_i}_2}$, $\hat{v}\equiv\Pi(\frac{v}{\norm{v}})$ is the projected prior vector, and $\lambda \in [0,1]$ controls the bias of the query $u_i$ towards the prior $\hat{v}$. Similarly, the optimal $\lambda$ is obtained by Eq. (\ref{apeq:optimal-lambda}). Afterwards, we insert the query vectors $u_i$ into Eq. (\ref{eq:rgf}) to estimate the gradient and conduct the PGD process in Eq. (\ref{eq:pgd}) to generate an adversarial example.

\subsection{Radial symmetry of the gradient estimation framework}
\label{app:rgf}

    Here, we justify why radial symmetry is crucial for the RGF framework. Based on the C\&W method~\cite{carlini2017towards}, the gradient can be expressed as the linear combination of the amplitude of changes for a set of orthogonal bases,
    \begin{equation}\small
    	\nabla L(x)= \frac{\partial L(x)}{\partial e_i} e_i,
    \end{equation}
    where $e_i$ represents a unit vector in the $i^\text{th}$ basis, since a sufficiently small region around input $x$ can be regarded as a (hyper)plane. While the C\&W method is computationally intensive, a more efficient way is to adopt the RGF method~\cite{tu2019autozoom} by randomly querying the vectors $u_i$ from the distribution $\mathcal{U}$, \ie,
    \begin{equation}\small
    	\nabla L(x) \approx  \sum_i^q\frac{\partial L(x)}{\partial u_i}u_i.
    \end{equation}
    Let $u_i \equiv\mu+\Delta_i$, where $\mu$ is the mean and $\Delta_i$ is the variance for the $i^\text{th}$ query vector. $\nabla L(x)$ can be obtained by
    \begin{equation}\small
    \begin{split}
        \sum_i^q\frac{\partial L(x)}{\partial u_i}u_i &= \sum_i^q\frac{\partial L(x)}{\partial (\mu + \Delta_i)}(\mu + \Delta_i) \\&= 
        \sum_i^q(\mu + \Delta_i)(\mu + \Delta_i)^T\nabla f \\&= q\mu^2 + \sum_i\Delta_i\Delta_i^T\nabla L(x).   
    \end{split}
    \end{equation}
    Thus, to adequately approximate the gradient $\nabla L(x)$, the mean $\mu$ is required to be a zero vector, and $\sum_i\Delta_i\Delta_i^T$ needs to be unbiased along any axis, \ie, the query distribution $\mathcal{U}$ has to be radially symmetric.

%% file: appendix/C_Jacobian.tex
\section{Evaluations of the self-guiding prior}
\label{app:prior}

    In the following, we first evaluate the effectiveness of the proposed self-guiding prior vector compared with the transfer-based prior vector~\cite{cheng2019improving}. Afterward, we present visual results of perturbing a single pixel to directly illustrate the sparse and diagonal properties of the Jacobian matrix of Img2Img GANs.

\subsection{Comparisons of self-guiding prior and transfer-based prior}
    \label{app-sub:prior-performance}
    \input{thefigs/app_prior_comparisons}
    Following~\cite{cheng2019improving}, we measure the correctness of a prior by cosine similarity $\alpha$ between the prior vector and the actual gradient. Transfer-based priors are acquired from surrogate models trained by the $100$ test samples with the same architectures and conditions as the threat models. Even though the transfer-based priors exploited surrogate models that are challenging to prepare, Figure \ref{fig:cosine} shows that the $\alpha$ values of the self-guiding priors are greater than that of transfer-based priors by at least $216\%$. The above result manifests the effectiveness of the proposed self-guiding prior.

\subsection{Visualizing the Jacobian matrix}
\label{app-sub:diagonality-jacob}
\input{thefigs/app_response}

    Here, for a random test sample for \textsc{black2blond}, we visualize the Jacobian matrix to further motivate the use of self-guiding priors. Recall that each element of a Jacobian matrix is defined by how each pixel affects itself (diagonal terms) and the other pixels (non-diagonal terms) under perturbation. To examine the Jacobian matrix of image translation functions, we perturb a pixel of a test image with a small value $h$ to find the response vector $\omega = \cfrac{1}{h}(\mathbb{T}(x_0 + h\hat{e_i}) - \mathbb{T}(x_0))$ in the output. Figure \ref{fig:response-example} shows the response where perturbations only result in a localized and sparse difference in the output. The localized and sparse responses indicate that the Jacobian matrix is sufficiently diagonal corresponding to the perturbed pixel.

%% file: thefigs/app_prior_comparisons.tex
\begin{figure}[ht]
    \centering
    %     \begin{subfigure}{.48\linewidth}
    % 		\includegraphics[width=\linewidth]{theimgs/app_Jacobians/prior_comparisons/str2segcompare.jpg}
    %         \caption{\textsc{str2seg}}
    %     \end{subfigure}
        
	\includegraphics[width=.8\linewidth]{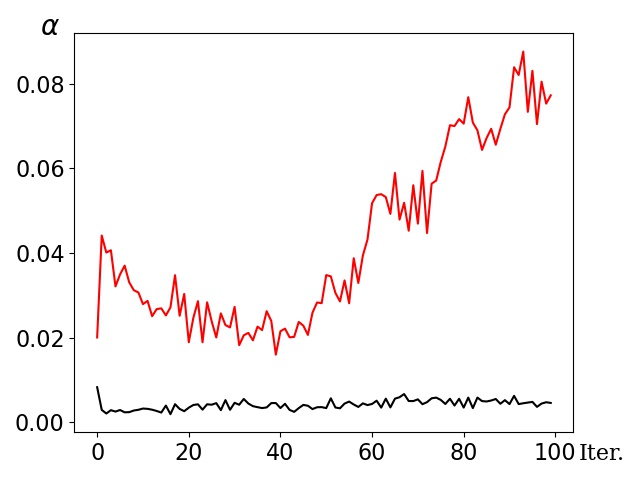}

    \caption{Averaged cosine similarity with true gradient value $\alpha$ for both priors measured on \textsc{black2blond} along the iterative process of white-box PGD attacks. (red: the \emph{self-guiding} prior; black: the \emph{transfer-based} prior.)}
    \label{fig:cosine}
    \end{figure}

%% file: thefigs/app_response.tex
\begin{figure}[h]
    \centering
    	\begin{subfigure}{\linewidth}
            \centering
    		\includegraphics[width=0.24\linewidth]{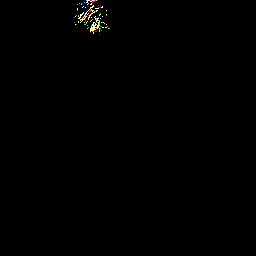}
    		\includegraphics[width=0.24\linewidth]{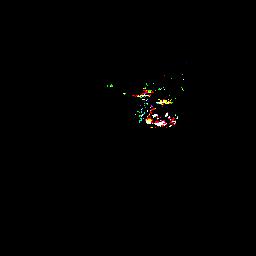}
            \centering
    		\includegraphics[width=0.24\linewidth]{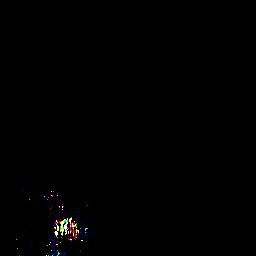}
    		\includegraphics[width=0.24\linewidth]{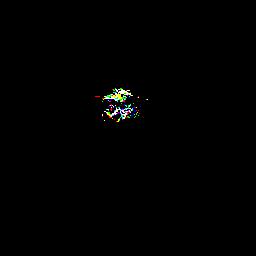}
    	\end{subfigure}
    \caption{Examples of typical responses when perturbing a single pixel in the input of model \textsc{black2blond}.}
    \label{fig:response-example}
    \end{figure}

%% file: appendix/D_distort.tex
\section{Distorting Attack}\label{appx:distort}
Following the definition for nullifying attack, one possible modification is to adjust the adversarial loss to obtain a loss suitable for the \emph{distorting attack}. 
 \begin{definition}
    \label{def:dist}
    \textbf{Distorting attack.} 
        The distorting attack aims to destroy the image translation process such that the adversarial example $x^*$ is mapped away from the legitimate target domain $Y$, which can be achieved by the distorting loss $L_\text{Dist} = (\norm{\mathbb{T}(x^*) - y_0}_2)^2, \,\,y_0 = \mathbb{T}(x_0)$, where $x_0, y_0$ are the original input and output of the image translation function, respectively.
    \end{definition}
    
\begin{figure}[t]
    \centering
	    \includegraphics[width=0.23\linewidth]{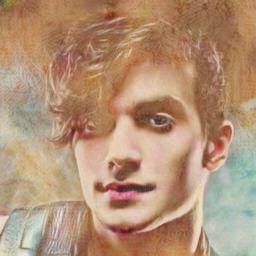}
	    \includegraphics[width=0.23\linewidth]{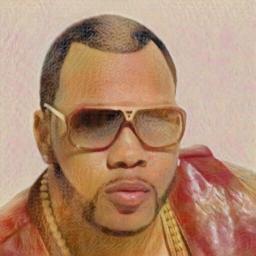}
	    \includegraphics[width=0.23\linewidth]{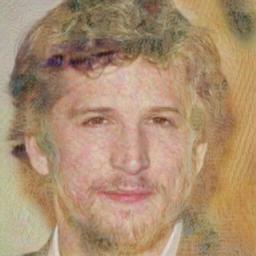}
	    \includegraphics[width=0.23\linewidth]{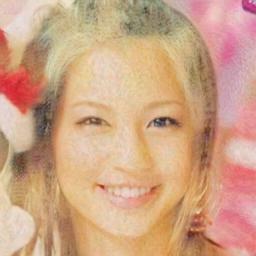}
	    \includegraphics[width=0.23\linewidth]{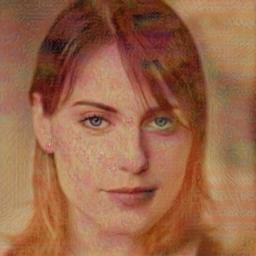}
	    \includegraphics[width=0.23\linewidth]{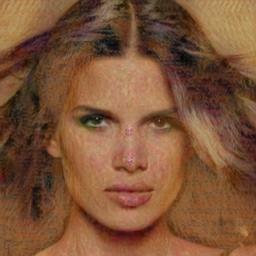}
	    \includegraphics[width=0.23\linewidth]{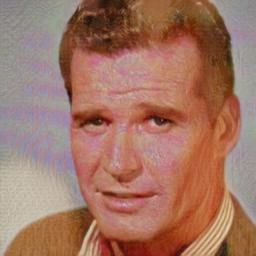}
	    \includegraphics[width=0.23\linewidth]{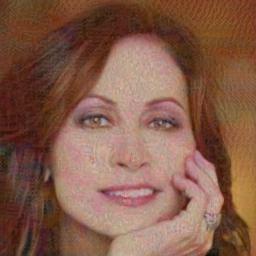}
    \caption{Distorting Attack example results. The above figures are created with white box attack following~\cite{yeh2020disrupting} for the model \textsc{black2blond}. In comparison to Figure~\ref{fig:ext-blond}, Distorting attack creates figures with \emph{blond} hairs, as well as other random colors, on the final portrait. On the contrary, Nullifying attack returns the image back to the original form.}
    \label{fig:dem}
    \end{figure}

While distorting attack provides an alternative, we display in Figure \ref{fig:dem} the result of Distorting attack on \textsc{black2blond}, with the same input images as Figure \ref{fig:ext-blond}. As can be seen, Distorting attack could not protect the images from the intended manipulation, \ie, causing the hair to become blond. Instead, it creates random distributed patches of blond and other colors throughout the portrait.

%% file: appendix/E_extended_results.tex
\section{Supplementary experimental results} \label{app:extensive}

    To further evaluate the visual quality of our \modelNameShort, we presents eight additional examples for each threat model, \ie, \textsc{black2blond} (in Figure \ref{fig:ext-blond}), \textsc{none2glasses} (in Figure \ref{fig:glasses}), and \textsc{blue2red} (in Figure \ref{fig:red2blue}), for the Nullifying Attack. Furthermore, we prepared $4$ additional Img2Img GANs~\cite{zhu2017unpaired}, including \textsc{str2seg} \textsc{str2seg-mapillery}, \textsc{facade2label}, and \textsc{night2day}, to demonstrate the generality of \modelNameShort for the \emph{Distorting Attack} scheme (Figures \ref{fig:str2seg-start} to \ref{fig:night2day}). 
    
    In the following (Figures \ref{fig:ext-blond} to \ref{fig:night2day}), ``input,'' ``expected,'' and ``adversarial'' columns display the input images, the Img2Img GAN outputs and the adversarial examples created by \modelNameShort, respectively, whereas the final ``distorting'' or ``nullifying'' columns display the final results under the distorting or nullifying attack.
    
\subsection{Qualitative results}
    
    We present additional qualitative results for model \textsc{black2blond}, \textsc{none2glasses}, and \textsc{blue2red}.\footnote{Similar to the notorious Img2Img GAN-based DeepFake (or DeepNude), these three models also transfer the input images into different styles while keeping the input semantics.} Figure \ref{fig:ext-blond} displays portraits of celebrities whose hair colors are changed to blond by \textsc{black2blond}. These portraits can be identified to be the same person with the black hair in the input image. We use these models as the substitute models to show how Img2Img GANs can be applied to manipulate pictures of targeted individuals and defame their identity. As presented in Figure \ref{fig:ext-blond}, \modelNameShort successfully creates adversarial examples that nullify \textsc{black2blond} and retain the blackness of the hair color. As the difference between the adversarial images and the original input images is indistinguishable by the human eyes, \modelNameShort successfully protects the pictures without pixelating the faces. 
    
    Figures \ref{fig:glasses} and \ref{fig:red2blue} show similar results. In Figure \ref{fig:glasses}, although insignificant patterns may remain on the nullifying result, the nullifying results clearly remove the pair of glasses that was added by \textsc{none2glasses} in the ``expected'' columns. Besides, in Figure \ref{fig:red2blue}, while a portion of cloths may remain dark purple on the nullifying results, the color tone of the results is much closer to the input photos of blue shirts compared to the translated portraits with bright red colors. Therefore,  it demonstrates the potential of \modelNameShort to \emph{defend} against the immoral modifications of DeepFake by applying the nullifying attack. 
    
    \input{thefigs/app_final_visuals/blond}
    \input{thefigs/app_final_visuals/bglasses}
    \input{thefigs/app_final_visuals/blue}

\subsection{Qualitative examples for \modelNameShort with the distorting attack}
   
   As explained in Appendix \ref{appx:distort}, the distorting attack scheme may be achieved by adjusting the nullifying loss (Definition \ref{def:null}) by the distorting loss (Definition \ref{def:dist}). In the following, we prepared three relevant Img2Img GANs as threat models to demonstrate our \modelNameShort under the distorting attack scheme. Each threat model is detailed as follows. 1) \textsc{str2seg} translates street scenes to semantic segmentation maps trained on the Cityscapes dataset~\cite{Cordts2016Cityscapes} (Figure \ref{fig:str2seg-start}). 2) \textsc{facade2label}, which translates facade images to label maps (Figure \ref{fig:facade2label})~\cite{zhu2017unpaired}. 
    3) \textsc{night2day}, which translates night (or foggy) street scenes to clear daytime street scenes, trained on the NightOwls dataset~\cite{neumann2018nightowls} and the Mapillery Vistas dataset (Figure \ref{fig:night2day}).
   
   In all three cases, \modelNameShort successfully distorts the image-to-image translation process. Figure \ref{fig:str2seg-start} shows that the purple region in the lower half (representing the road area) is sporadically replaced by the pink (pedestrian sidewalk) and black (the car outline) color after distorting.
   Similarly, in Figure \ref{fig:facade2label}, \modelNameShort disrupts the semantic labeling of \textsc{facades2labels}, \eg, causing the red blocks in the corners (representing background) of the expected results to be replaced with cyan blocks (representing windows). Last, in Figure \ref{fig:night2day}, \modelNameShort causes \textsc{night2day} to be blurred with white or black blobs in the final output. While the results are successful, notice that all three results for distorting attacks are drastically \emph{different} and not as consistent as the nullifying attack.

    \input{thefigs/app_final_visuals/str2seg}
    \input{thefigs/app_final_visuals/facade2lable}
    \input{thefigs/app_final_visuals/night}
    

%% file: thefigs/app_final_visuals/blond.tex
\begin{figure*}[t]
    \centering
    	\begin{subfigure}{.12\linewidth}
            \centering
            \includegraphics[width=0.98\linewidth]{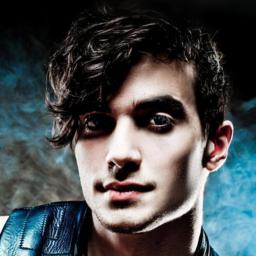}
            \includegraphics[width=0.98\linewidth]{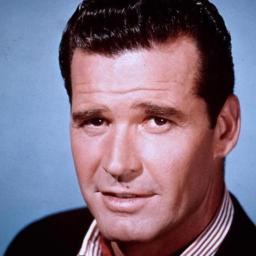}
            \includegraphics[width=0.98\linewidth]{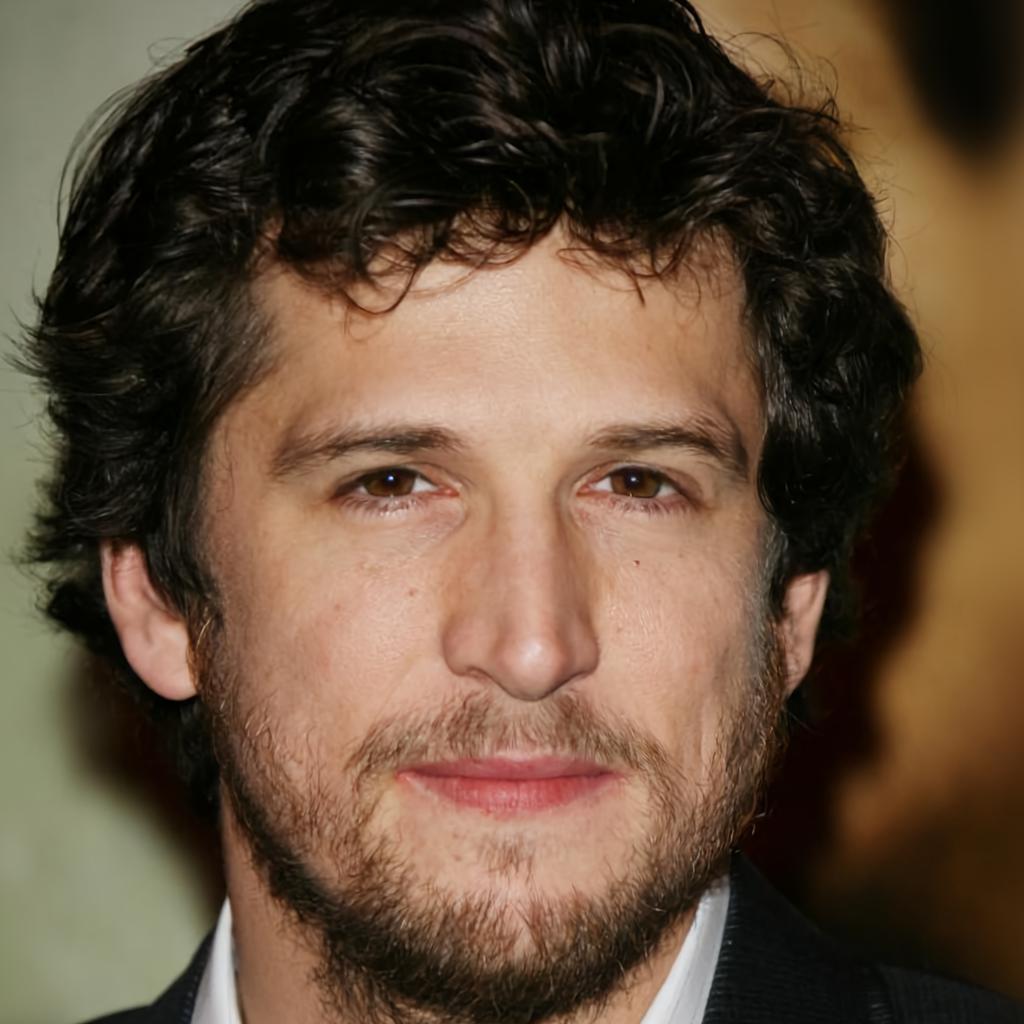}
    		\stackunder[5pt]{\includegraphics[width=0.98\linewidth]{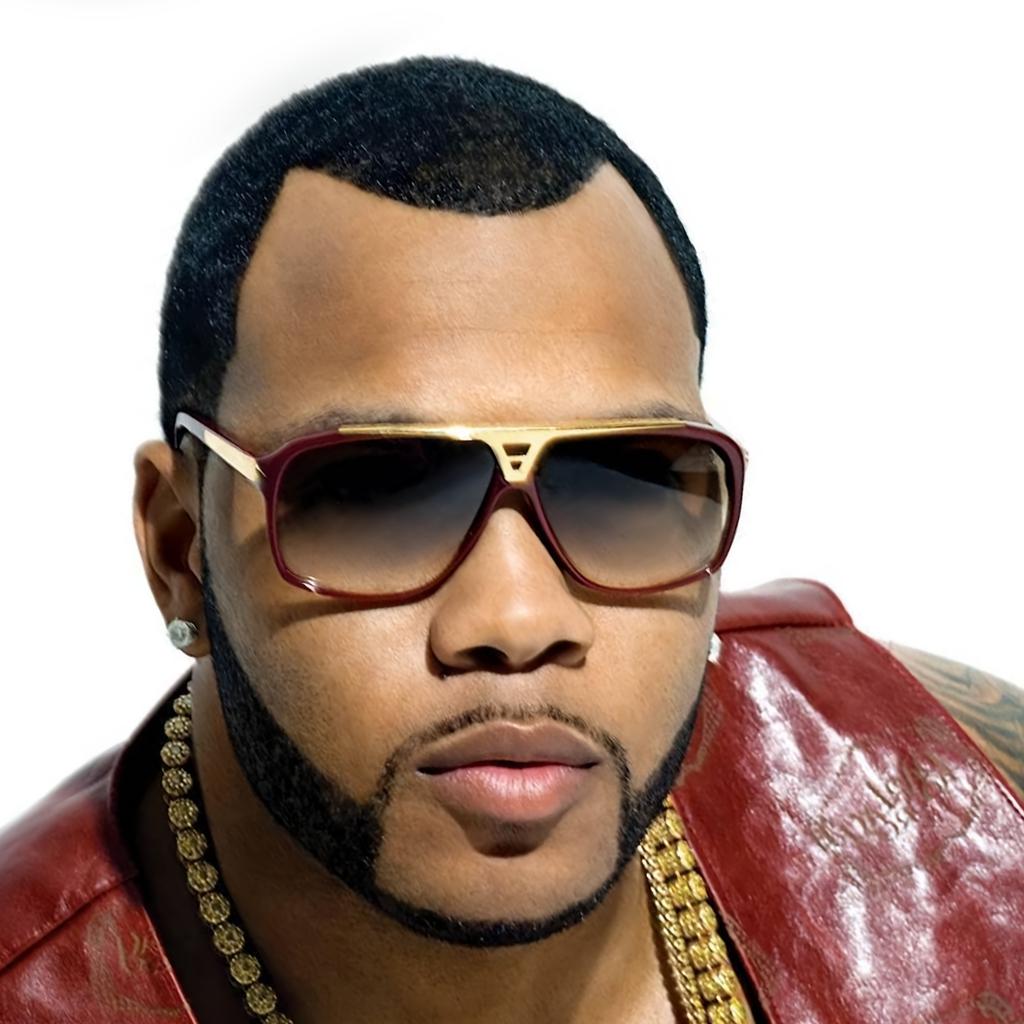}}{input}
    	\end{subfigure}
    	    \begin{subfigure}{.12\linewidth}
            \centering
            \includegraphics[width=0.98\linewidth]{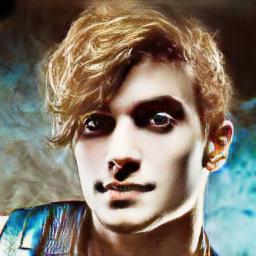}
            \includegraphics[width=0.98\linewidth]{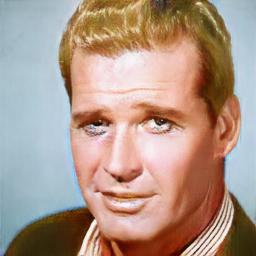}
            \includegraphics[width=0.98\linewidth]{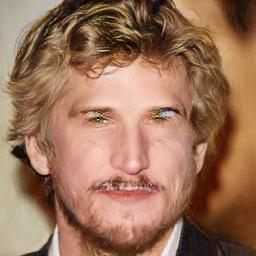}
            \stackunder[5pt]{\includegraphics[width=0.98\linewidth]{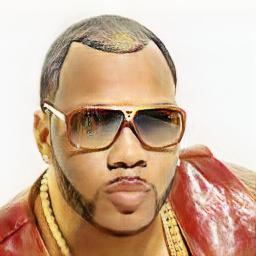}}{expected}
    	\end{subfigure}
        \begin{subfigure}{.12\linewidth}
            \centering
            \includegraphics[width=0.98\linewidth]{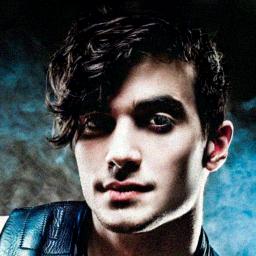}
            \includegraphics[width=0.98\linewidth]{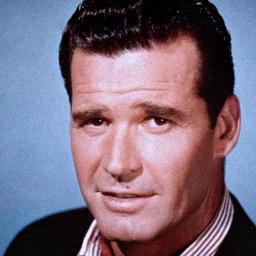}
            \includegraphics[width=0.98\linewidth]{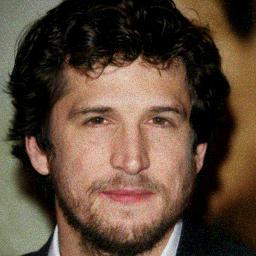}
            \stackunder[5pt]{\includegraphics[width=0.98\linewidth]{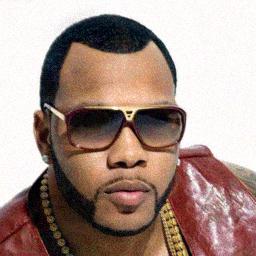}}{adversarial}
    	\end{subfigure}	
        \begin{subfigure}{.12\linewidth}
            \centering
            \includegraphics[width=0.98\linewidth]{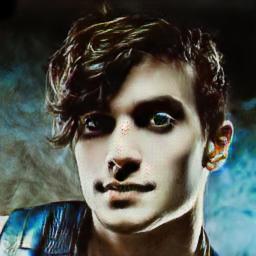}
            \includegraphics[width=0.98\linewidth]{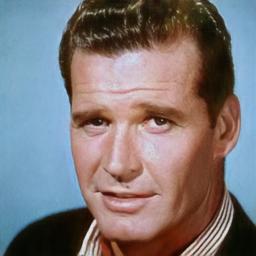}
            \includegraphics[width=0.98\linewidth]{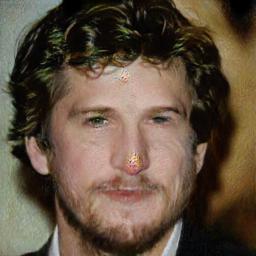}
            \stackunder[5pt]{\includegraphics[width=0.98\linewidth]{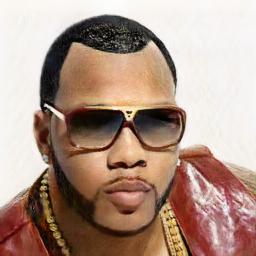}}{nullifying}
        \end{subfigure}
        \begin{subfigure}{.12\linewidth}
            \centering
            \includegraphics[width=0.98\linewidth]{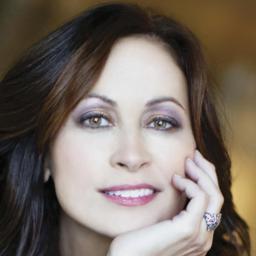}
    		\includegraphics[width=0.98\linewidth]{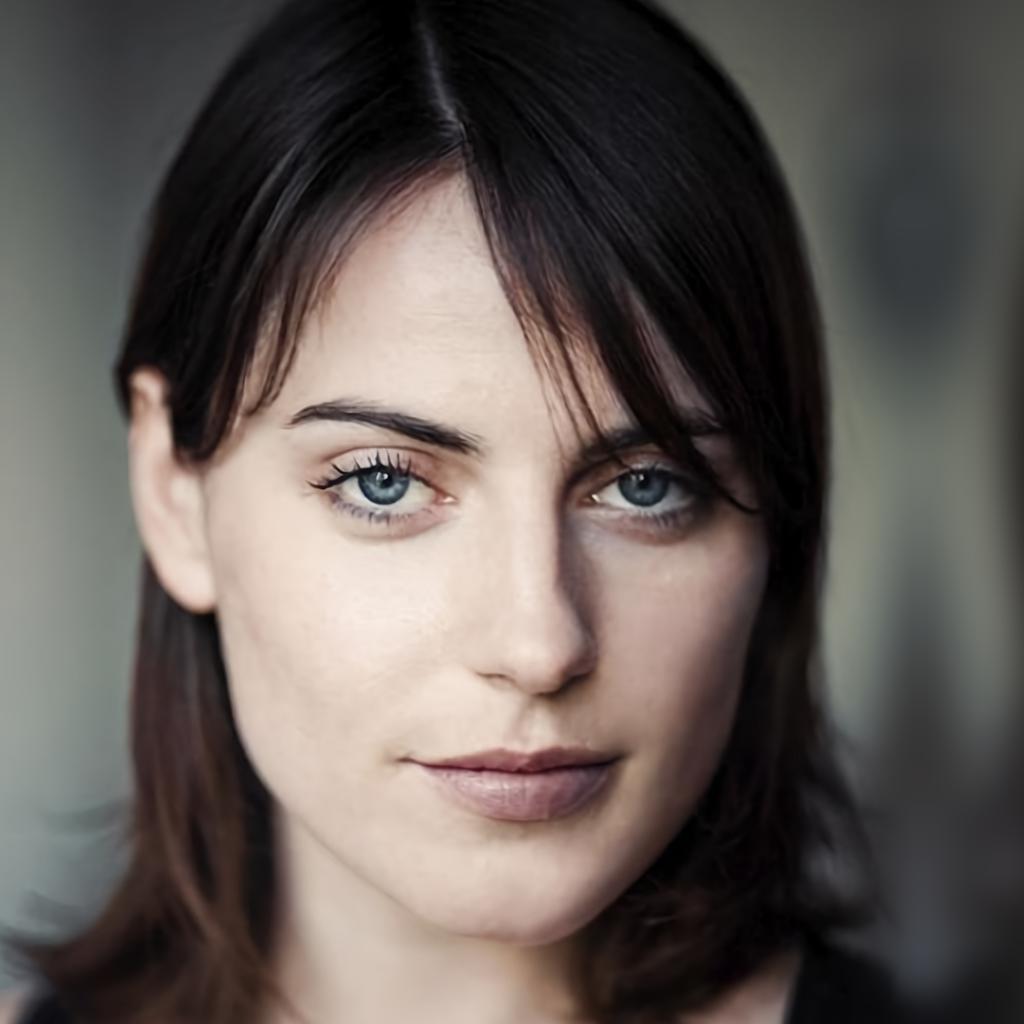}
    		\includegraphics[width=0.98\linewidth]{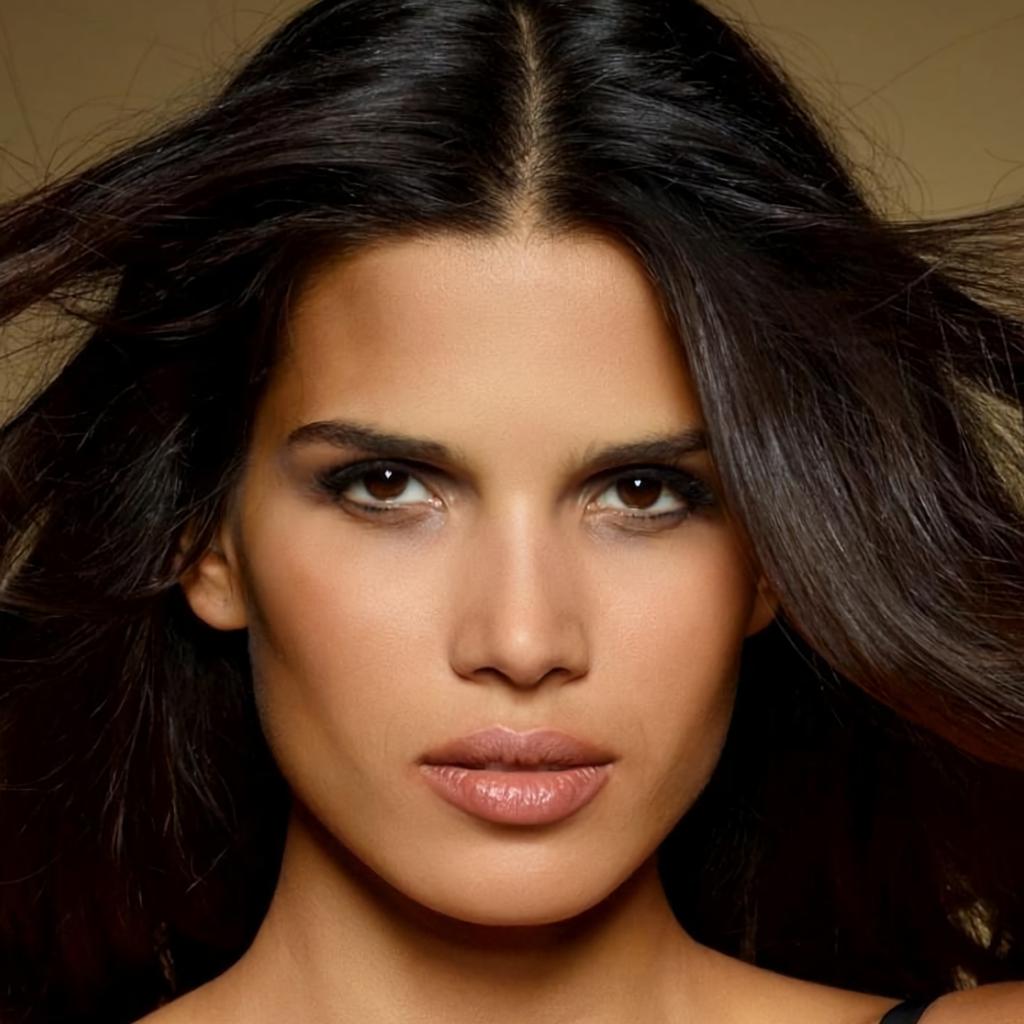} 
            \stackunder[5pt]{\includegraphics[width=0.98\linewidth]{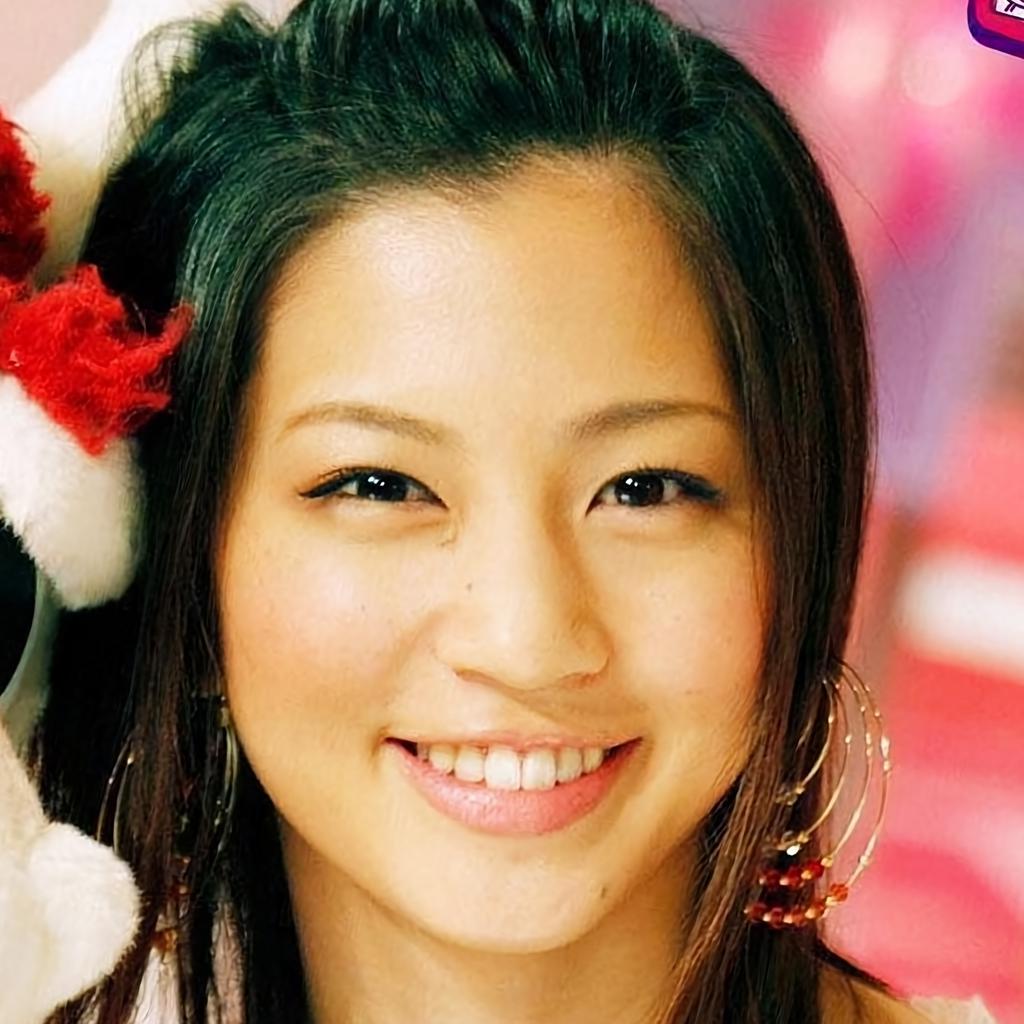}}{input}
        \end{subfigure}
        \begin{subfigure}{.12\linewidth}
            \centering
            \includegraphics[width=0.98\linewidth]{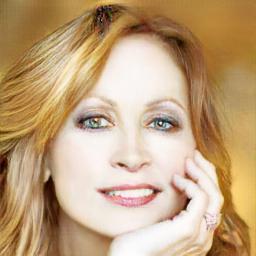}
            \includegraphics[width=0.98\linewidth]{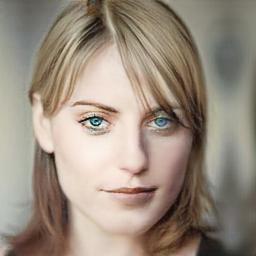}
            \includegraphics[width=0.98\linewidth]{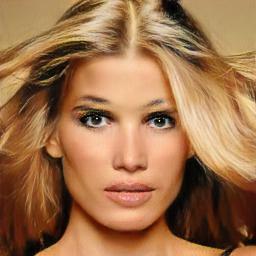}
            \stackunder[5pt]{\includegraphics[width=0.98\linewidth]{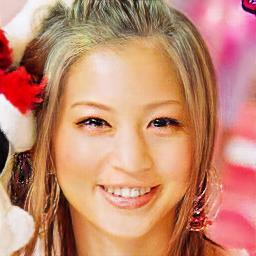}}{expected}
        \end{subfigure}
        \begin{subfigure}{.12\linewidth}
            \centering
            \includegraphics[width=0.98\linewidth]{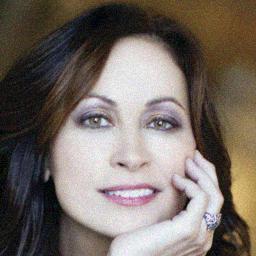}
            \includegraphics[width=0.98\linewidth]{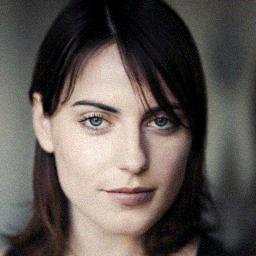}
            \includegraphics[width=0.98\linewidth]{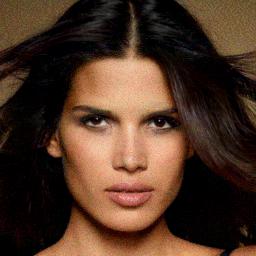}
            \stackunder[5pt]{\includegraphics[width=0.98\linewidth]{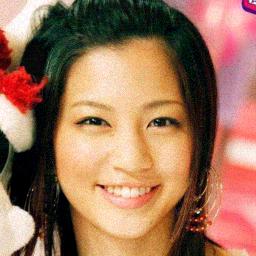}}{adversarial}
        \end{subfigure}
        \begin{subfigure}{.12\linewidth}
            \centering
            \includegraphics[width=0.98\linewidth]{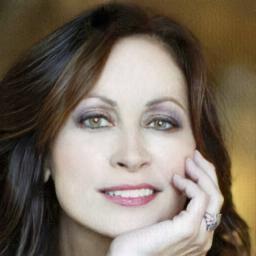}
            \includegraphics[width=0.98\linewidth]{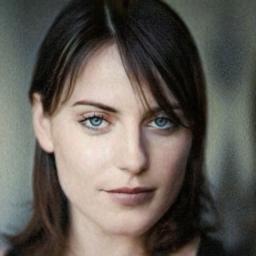}
            \includegraphics[width=0.98\linewidth]{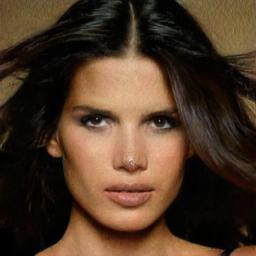}
            \stackunder[5pt]{\includegraphics[width=0.98\linewidth]{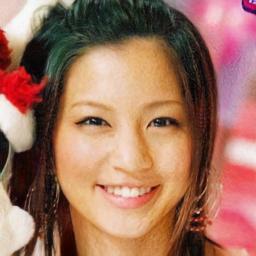}}{nullifying}
        \end{subfigure}
    \caption{Qualitative results of \modelNameShort against \textsc{black2blond} under the nullifying attack scheme. Even though \textsc{black2blond} works perfectly and creates
    blond-hair portraits from black-hair portraits, \modelNameShort adds imperceptible adversarial perturbations to create adversarial examples that cause \textsc{black2blond} to return images similar to the original input images. Namely, the ``nullifying'' results display black-hair portraits instead of blond-hair portraits.}
    \label{fig:ext-blond}
    \end{figure*}

%% file: thefigs/app_final_visuals/bglasses.tex
\begin{figure*}[t]
    \centering
	\begin{subfigure}{.12\linewidth}
        \centering
        \includegraphics[width=0.98\linewidth]{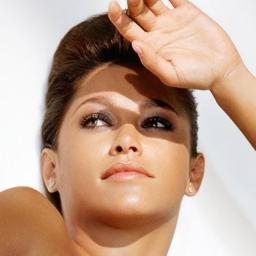}
	    \includegraphics[width=0.98\linewidth]{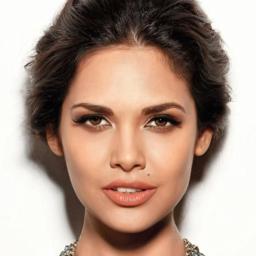}
		\includegraphics[width=0.98\linewidth]{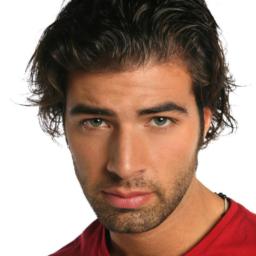}
		\stackunder[5pt]{\includegraphics[width=0.98\linewidth]{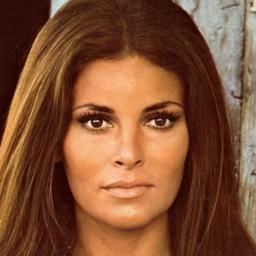}}{input}
	\end{subfigure}
	    \begin{subfigure}{.12\linewidth}
        \centering
        \includegraphics[width=0.98\linewidth]{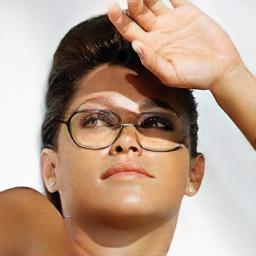}
        \includegraphics[width=0.98\linewidth]{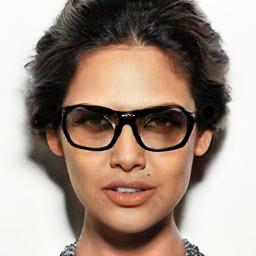}
        \includegraphics[width=0.98\linewidth]{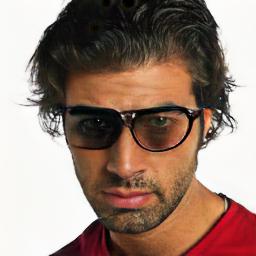}
        \stackunder[5pt]{\includegraphics[width=0.98\linewidth]{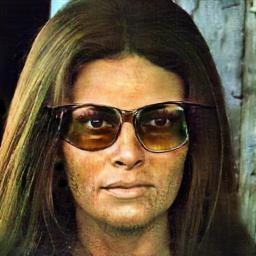}}{expected}
	\end{subfigure}
    \begin{subfigure}{.12\linewidth}
        \centering
        \includegraphics[width=0.98\linewidth]{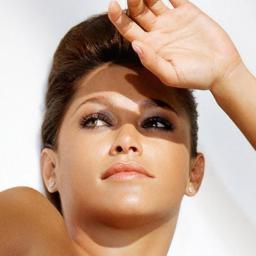}
        \includegraphics[width=0.98\linewidth]{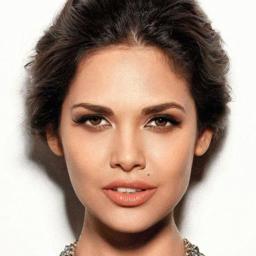}
        \includegraphics[width=0.98\linewidth]{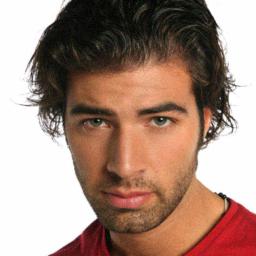}
        \stackunder[5pt]{\includegraphics[width=0.98\linewidth]{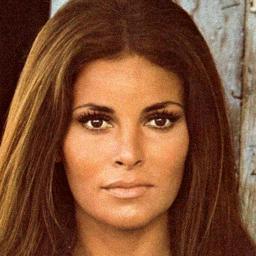}}{adversarial}
	\end{subfigure}	
    \begin{subfigure}{.12\linewidth}
        \centering
        \includegraphics[width=0.98\linewidth]{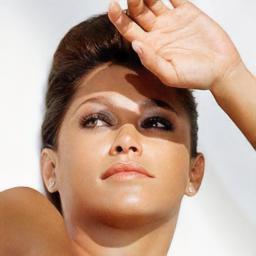}
        \includegraphics[width=0.98\linewidth]{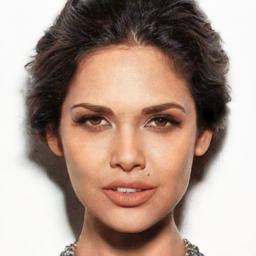}
        \includegraphics[width=0.98\linewidth]{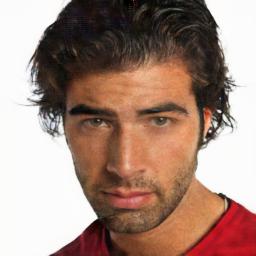}
        \stackunder[5pt]{\includegraphics[width=0.98\linewidth]{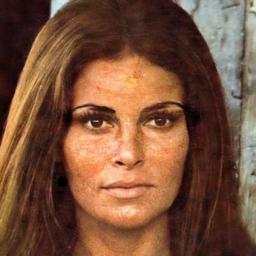}}{nullifying}
    \end{subfigure}
    \begin{subfigure}{.12\linewidth}
        \centering
        \includegraphics[width=0.98\linewidth]{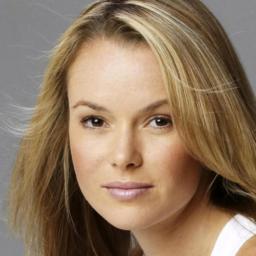}
		\includegraphics[width=0.98\linewidth]{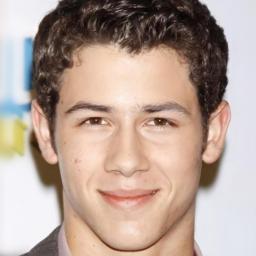}
		\includegraphics[width=0.98\linewidth]{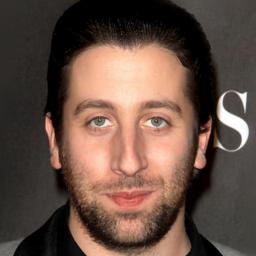} 
        \stackunder[5pt]{\includegraphics[width=0.98\linewidth]{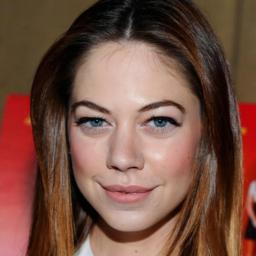}}{input}
	\end{subfigure}
	    \begin{subfigure}{.12\linewidth}
        \centering
        \includegraphics[width=0.98\linewidth]{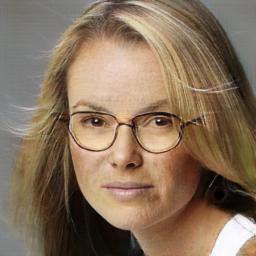}
        \includegraphics[width=0.98\linewidth]{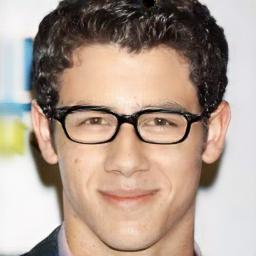}
        \includegraphics[width=0.98\linewidth]{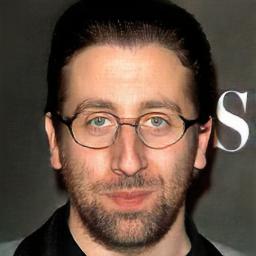}
        \stackunder[5pt]{\includegraphics[width=0.98\linewidth]{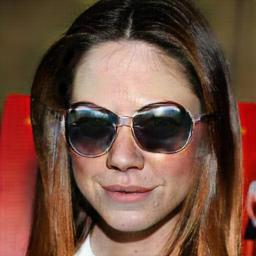}}{expected}
	\end{subfigure}
    \begin{subfigure}{.12\linewidth}
        \centering
        \includegraphics[width=0.98\linewidth]{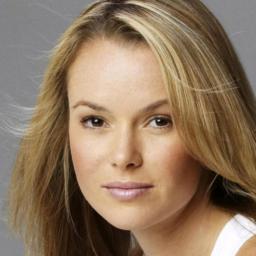}
        \includegraphics[width=0.98\linewidth]{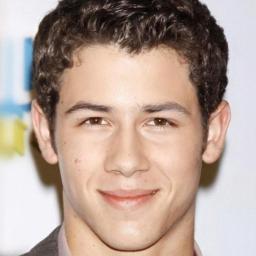}
        \includegraphics[width=0.98\linewidth]{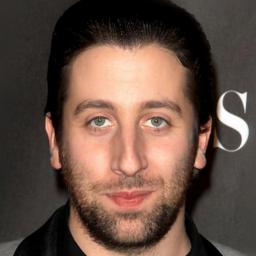}
        \stackunder[5pt]{\includegraphics[width=0.98\linewidth]{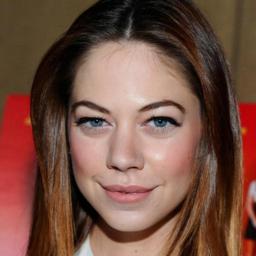}}{adversarial}
	\end{subfigure}	
    \begin{subfigure}{.12\linewidth}
        \centering
        \includegraphics[width=0.98\linewidth]{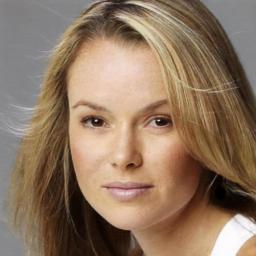}
        \includegraphics[width=0.98\linewidth]{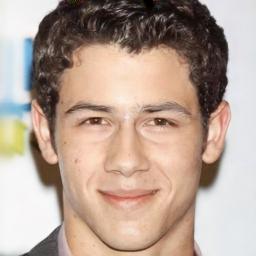}
        \includegraphics[width=0.98\linewidth]{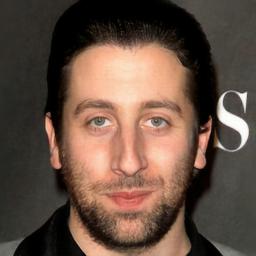}
        \stackunder[5pt]{\includegraphics[width=0.98\linewidth]{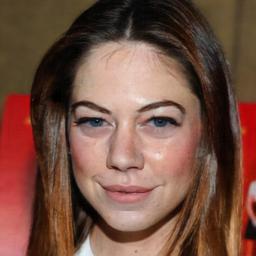}}{nullifying}
    \end{subfigure}
    \caption{Qualitative results of \modelNameShort against \textsc{none2glasses} under the nullifying attack scheme. The ``expected'' columns displays that the outputs of \textsc{none2glasses} clearly adds eyeglasses to each portraits in the corresponding ``input'' columns. Nonetheless, by adding human-imperceptible modifications (the ``adversarial'' columns), \modelNameShort nullifies the model functionality and cause it to output the results in the ``nullifying'' columns, in which the original portraits are restored.}
    \label{fig:glasses}
\end{figure*}

%% file: thefigs/app_final_visuals/blue.tex
\begin{figure*}[t]
    \centering
	\begin{subfigure}{.12\linewidth}
        \centering
        \includegraphics[width=0.98\linewidth]{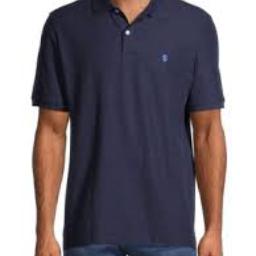}
	    \includegraphics[width=0.98\linewidth]{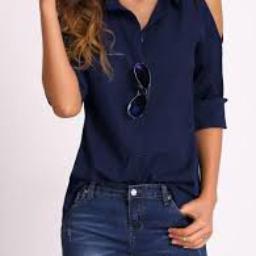}
		\includegraphics[width=0.98\linewidth]{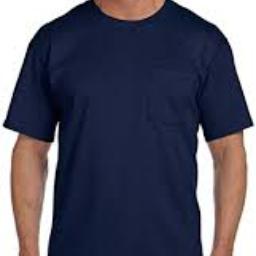}
		\stackunder[5pt]{\includegraphics[width=0.98\linewidth]{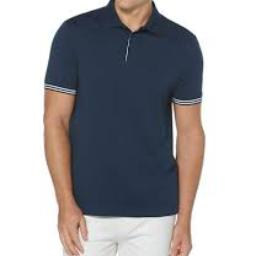}}{input}
	\end{subfigure}
	    \begin{subfigure}{.12\linewidth}
        \centering
        \includegraphics[width=0.98\linewidth]{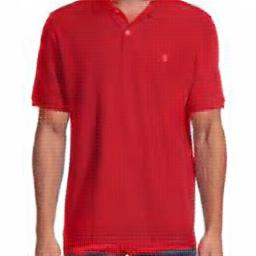}
        \includegraphics[width=0.98\linewidth]{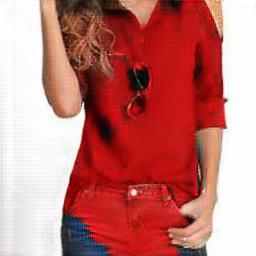}
        \includegraphics[width=0.98\linewidth]{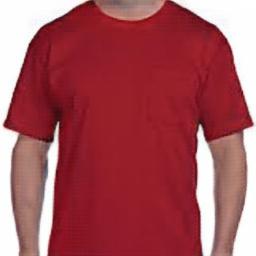}
        \stackunder[5pt]{\includegraphics[width=0.98\linewidth]{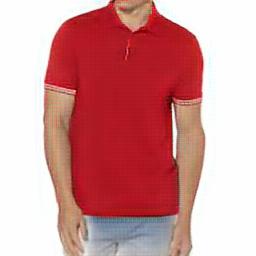}}{expected}
	\end{subfigure}
    \begin{subfigure}{.12\linewidth}
        \centering
        \includegraphics[width=0.98\linewidth]{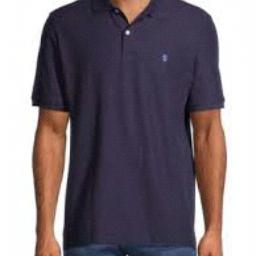}
        \includegraphics[width=0.98\linewidth]{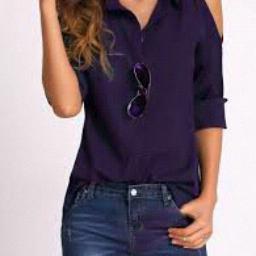}
        \includegraphics[width=0.98\linewidth]{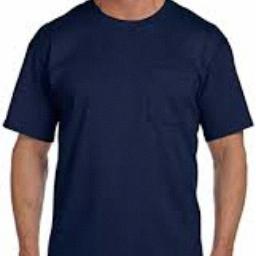}
        \stackunder[5pt]{\includegraphics[width=0.98\linewidth]{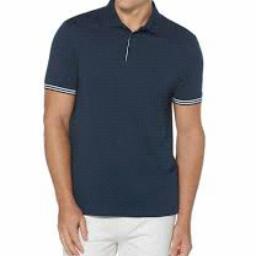}}{adversarial}
	\end{subfigure}	
    \begin{subfigure}{.12\linewidth}
        \centering
        \includegraphics[width=0.98\linewidth]{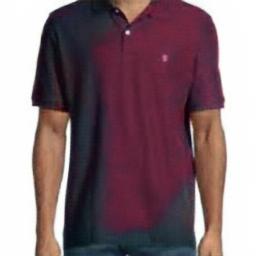}
        \includegraphics[width=0.98\linewidth]{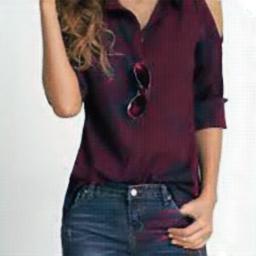}
        \includegraphics[width=0.98\linewidth]{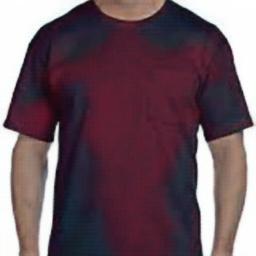}
        \stackunder[5pt]{\includegraphics[width=0.98\linewidth]{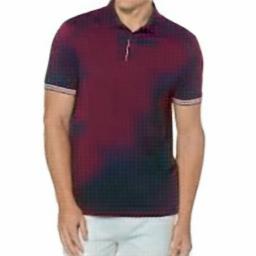}}{nullifying}
    \end{subfigure}
        	\begin{subfigure}{.12\linewidth}
        \centering
        \includegraphics[width=0.98\linewidth]{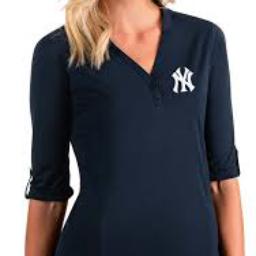}
		\includegraphics[width=0.98\linewidth]{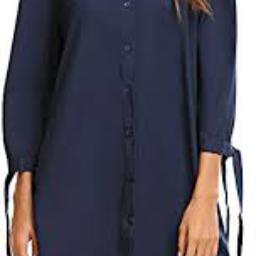}
		\includegraphics[width=0.98\linewidth]{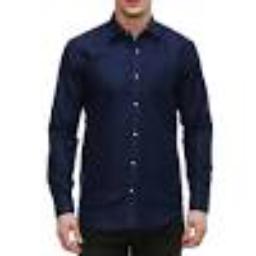} 
        \stackunder[5pt]{\includegraphics[width=0.98\linewidth]{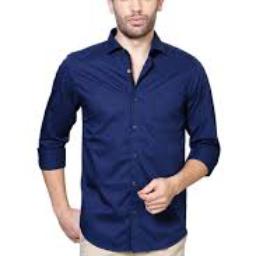}}{input}
	\end{subfigure}
	    \begin{subfigure}{.12\linewidth}
        \centering
        \includegraphics[width=0.98\linewidth]{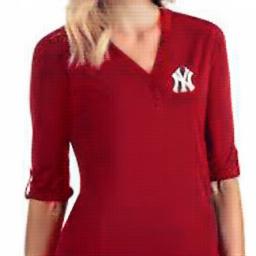}
        \includegraphics[width=0.98\linewidth]{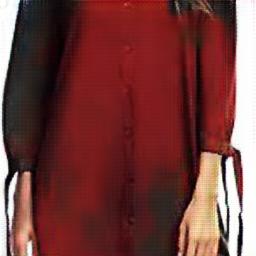}
        \includegraphics[width=0.98\linewidth]{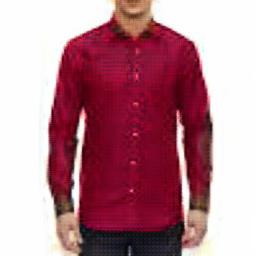}
        \stackunder[5pt]{\includegraphics[width=0.98\linewidth]{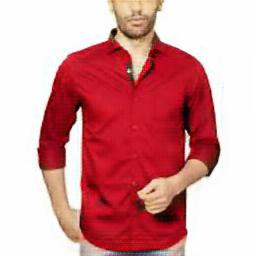}}{expected}
	\end{subfigure}
    \begin{subfigure}{.12\linewidth}
        \centering
        \includegraphics[width=0.98\linewidth]{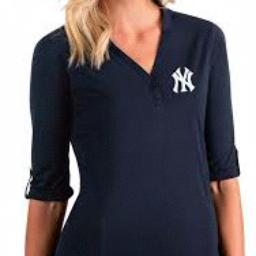}
        \includegraphics[width=0.98\linewidth]{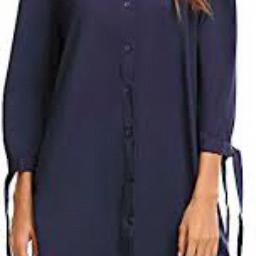}
        \includegraphics[width=0.98\linewidth]{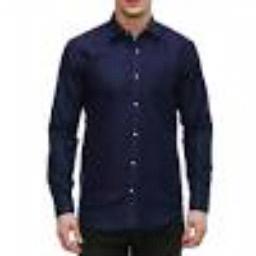}
        \stackunder[5pt]{\includegraphics[width=0.98\linewidth]{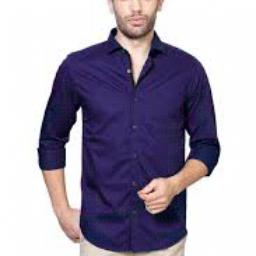}}{adversarial}
	\end{subfigure}	
    \begin{subfigure}{.12\linewidth}
        \centering
        \includegraphics[width=0.98\linewidth]{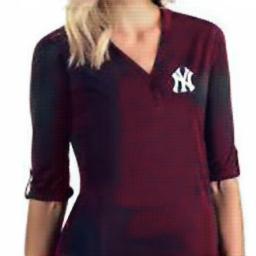}
        \includegraphics[width=0.98\linewidth]{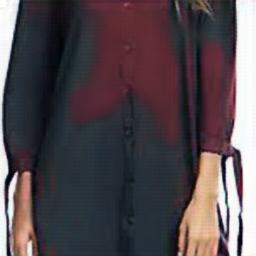}
        \includegraphics[width=0.98\linewidth]{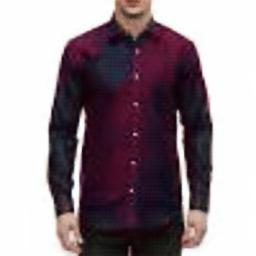}
        \stackunder[5pt]{\includegraphics[width=0.98\linewidth]{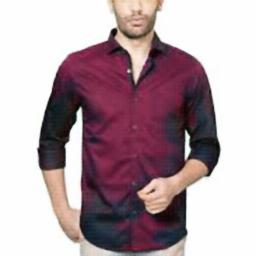}}{nullifying}
    \end{subfigure}
    \caption{Qualitative results of \modelNameShort against \textsc{blue2red} under the nullifying attack scheme. Although \modelNameShort couldn't completely retrieve the blue color in some portion of the resulting image, the color (\ie dark purple) is always much closer to the original color (\ie dark blue) than to the expected threat model output (\ie bright red).}
    \label{fig:red2blue}
\end{figure*}

%% file: thefigs/app_final_visuals/str2seg.tex
\begin{figure*}[t]
    \centering
    	\begin{subfigure}{.12\linewidth}
            \centering
            \includegraphics[width=0.98\linewidth]{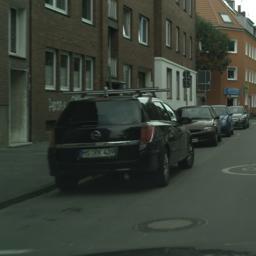}
            \includegraphics[width=0.98\linewidth]{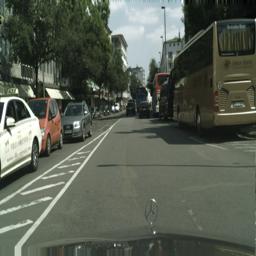}
            \includegraphics[width=0.98\linewidth]{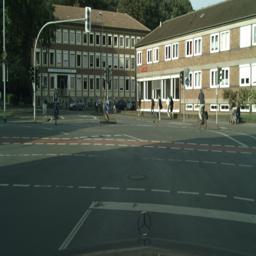}
            \stackunder[5pt]{\includegraphics[width=0.98\linewidth]{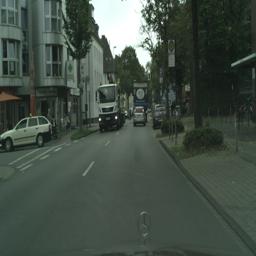}}{input}
    	\end{subfigure}
    	    \begin{subfigure}{.12\linewidth}
            \centering
            \includegraphics[width=0.98\linewidth]{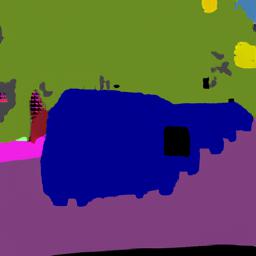}
            \includegraphics[width=0.98\linewidth]{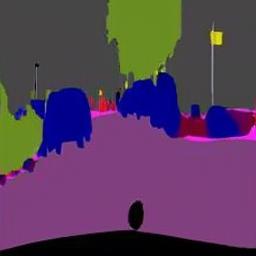}
            \includegraphics[width=0.98\linewidth]{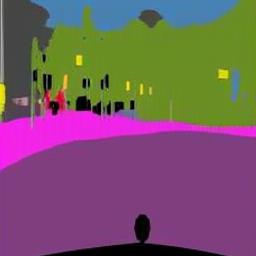}
            \stackunder[5pt]{\includegraphics[width=0.98\linewidth]{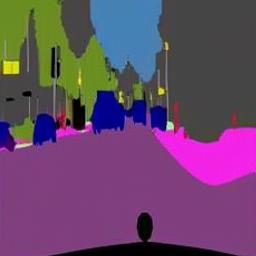}}{expected}
    	\end{subfigure}
        \begin{subfigure}{.12\linewidth}
            \centering
            \includegraphics[width=0.98\linewidth]{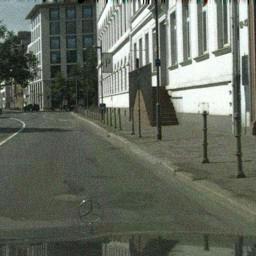}
            \includegraphics[width=0.98\linewidth]{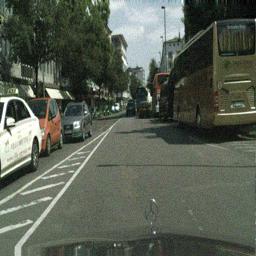}
            \includegraphics[width=0.98\linewidth]{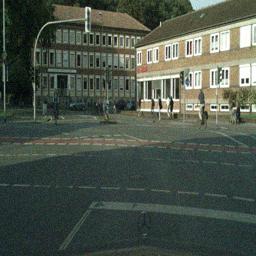}
            \stackunder[5pt]{\includegraphics[width=0.98\linewidth]{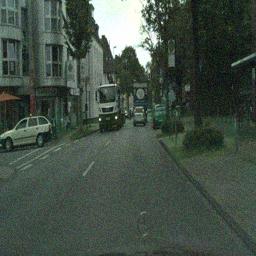}}{adversarial}
    	\end{subfigure}	
        \begin{subfigure}{.12\linewidth}
            \centering
            \includegraphics[width=0.98\linewidth]{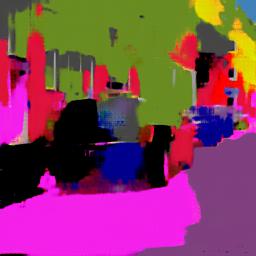}
            \includegraphics[width=0.98\linewidth]{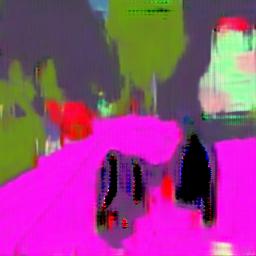}
            \includegraphics[width=0.98\linewidth]{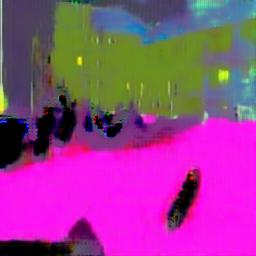}
            \stackunder[5pt]{\includegraphics[width=0.98\linewidth]{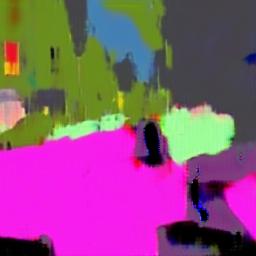}}{distorting}
        \end{subfigure}
        \begin{subfigure}{.12\linewidth}
            \centering
            \includegraphics[width=0.98\linewidth]{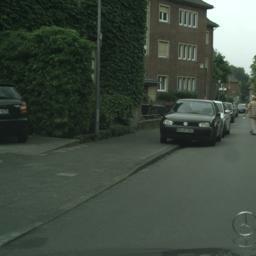}
            \includegraphics[width=0.98\linewidth]{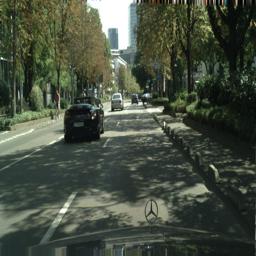}
            \includegraphics[width=0.98\linewidth]{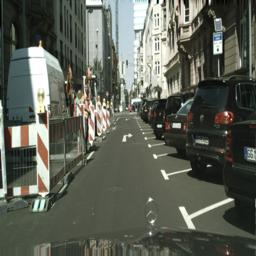}
            \stackunder[5pt]{\includegraphics[width=0.98\linewidth]{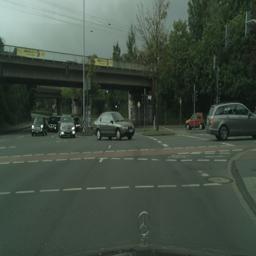}}{input}
    	\end{subfigure}
    	    \begin{subfigure}{.12\linewidth}
            \centering
            \includegraphics[width=0.98\linewidth]{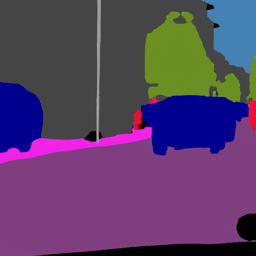}
            \includegraphics[width=0.98\linewidth]{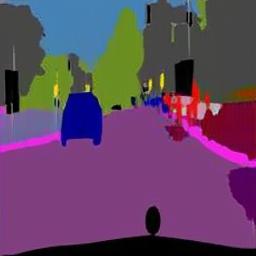}
            \includegraphics[width=0.98\linewidth]{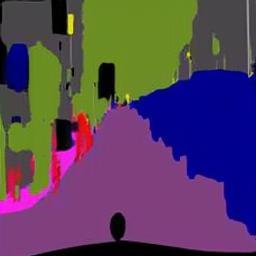}
            \stackunder[5pt]{\includegraphics[width=0.98\linewidth]{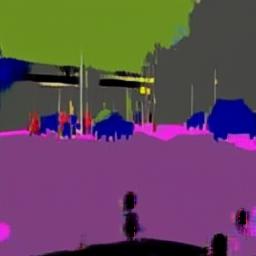}}{expected}
    	\end{subfigure}
        \begin{subfigure}{.12\linewidth}
            \centering
            \includegraphics[width=0.98\linewidth]{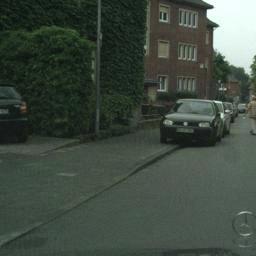}
            \includegraphics[width=0.98\linewidth]{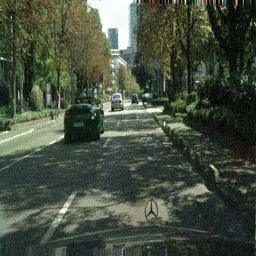}
            \includegraphics[width=0.98\linewidth]{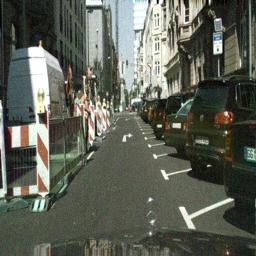}
            \stackunder[5pt]{\includegraphics[width=0.98\linewidth]{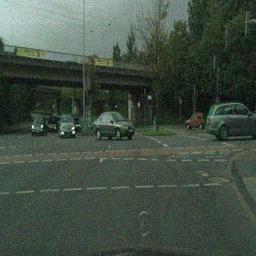}}{adversarial}
    	\end{subfigure}	
        \begin{subfigure}{.12\linewidth}
            \centering
            \includegraphics[width=0.98\linewidth]{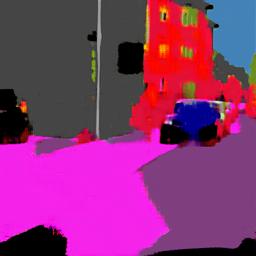}
            \includegraphics[width=0.98\linewidth]{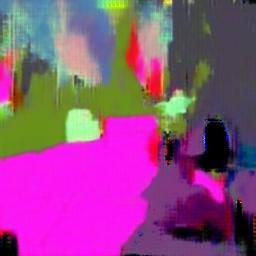}
            \includegraphics[width=0.98\linewidth]{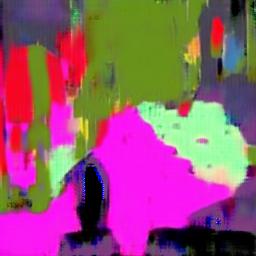}
            \stackunder[5pt]{\includegraphics[width=0.98\linewidth]{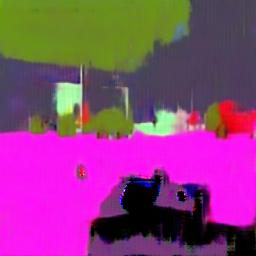}}{distorting}
        \end{subfigure}
    \caption{Qualitative results of \modelNameShort against \textsc{str2seg} under the distorting attack scheme. As displayed in the ``distorting'' columns, the configurations of color blocks are distorted while new colors also appear in irregular patterns (\eg, lime green in the upper parts and light pink in the lower parts of the distorting results). In particular, the adversarial attack result of the ``distorting'' images may fool self-driving applications into perceiving obstacles instead of roads (labeled by purple).}
    \label{fig:str2seg-start}
\end{figure*}

%% file: thefigs/app_final_visuals/facade2lable.tex
\begin{figure*}[t]
    \centering
    	\begin{subfigure}{.12\linewidth}
            \centering
            \includegraphics[width=0.98\linewidth]{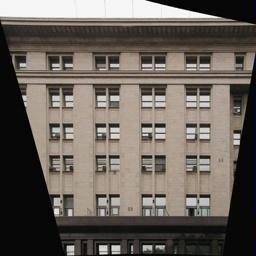}
            \includegraphics[width=0.98\linewidth]{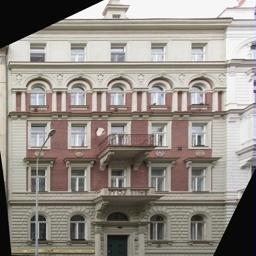}
            \includegraphics[width=0.98\linewidth]{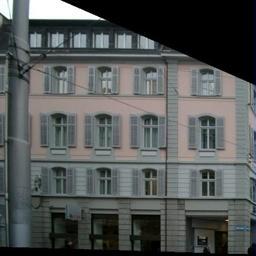}
            \stackunder[5pt]{\includegraphics[width=0.98\linewidth]{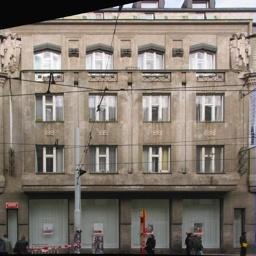}}{input}
    	\end{subfigure}
    	    \begin{subfigure}{.12\linewidth}
            \centering
            \includegraphics[width=0.98\linewidth]{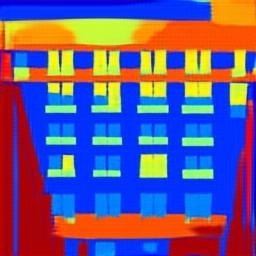}
            \includegraphics[width=0.98\linewidth]{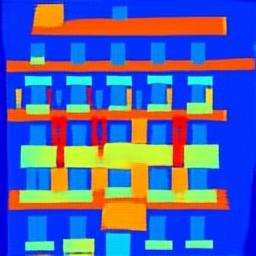}
            \includegraphics[width=0.98\linewidth]{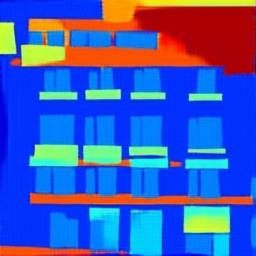}
            \stackunder[5pt]{\includegraphics[width=0.98\linewidth]{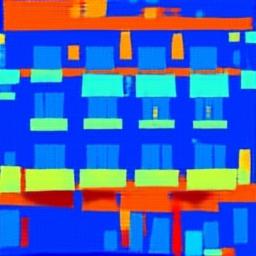}}{expected}
    	\end{subfigure}
        \begin{subfigure}{.12\linewidth}
            \centering
            \includegraphics[width=0.98\linewidth]{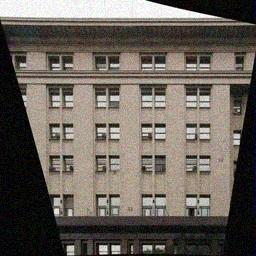}
            \includegraphics[width=0.98\linewidth]{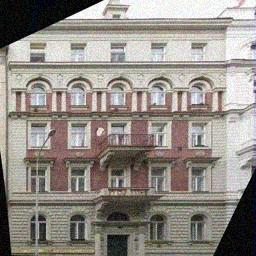}
            \includegraphics[width=0.98\linewidth]{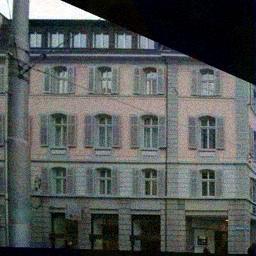}
            \stackunder[5pt]{\includegraphics[width=0.98\linewidth]{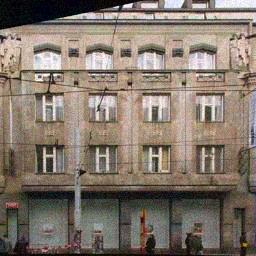}}{adversarial}
    	\end{subfigure}	
        \begin{subfigure}{.12\linewidth}
            \centering
            \includegraphics[width=0.98\linewidth]{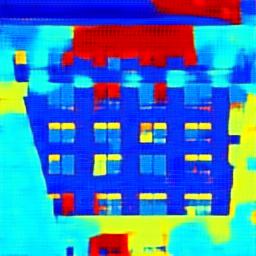}
            \includegraphics[width=0.98\linewidth]{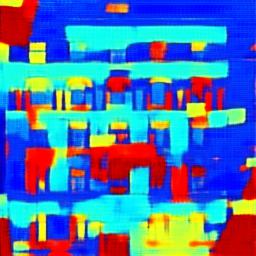}
            \includegraphics[width=0.98\linewidth]{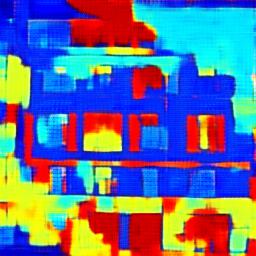}
            \stackunder[5pt]{\includegraphics[width=0.98\linewidth]{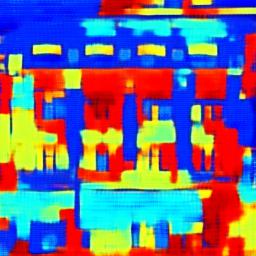}}{distorting}
        \end{subfigure}
        \begin{subfigure}{.12\linewidth}
            \centering
            \includegraphics[width=0.98\linewidth]{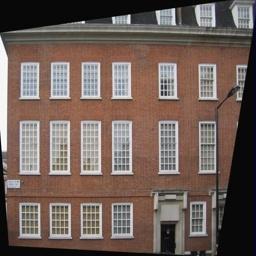}
            \includegraphics[width=0.98\linewidth]{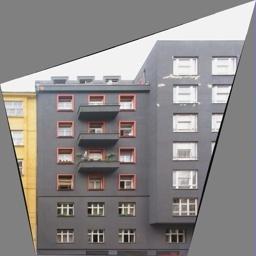}
            \includegraphics[width=0.98\linewidth]{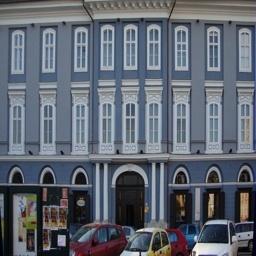}
            \stackunder[5pt]{\includegraphics[width=0.98\linewidth]{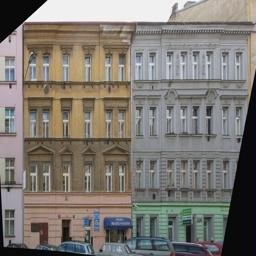}}{input}
    	\end{subfigure}
    	    \begin{subfigure}{.12\linewidth}
            \centering
            \includegraphics[width=0.98\linewidth]{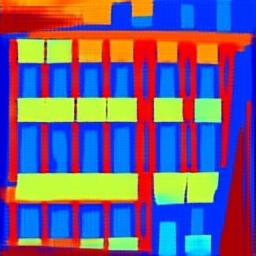}
            \includegraphics[width=0.98\linewidth]{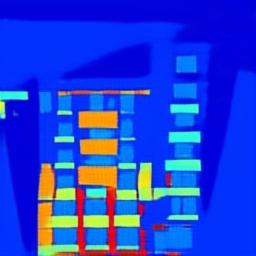}
            \includegraphics[width=0.98\linewidth]{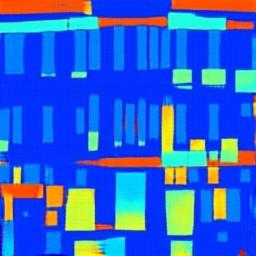}
            \stackunder[5pt]{\includegraphics[width=0.98\linewidth]{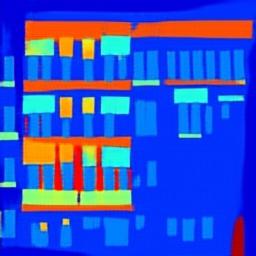}}{expected}
    	\end{subfigure}
        \begin{subfigure}{.12\linewidth}
            \centering
            \includegraphics[width=0.98\linewidth]{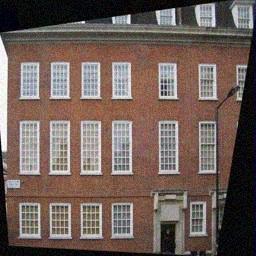}
            \includegraphics[width=0.98\linewidth]{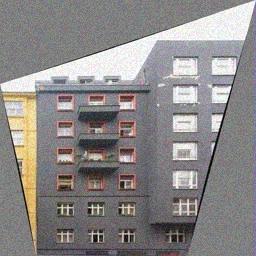}
            \includegraphics[width=0.98\linewidth]{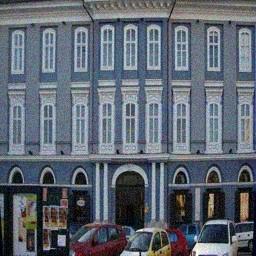}
            \stackunder[5pt]{\includegraphics[width=0.98\linewidth]{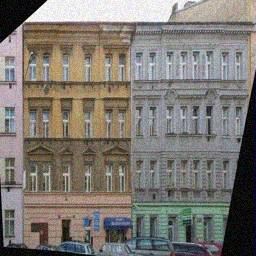}}{adversarial}
    	\end{subfigure}	
        \begin{subfigure}{.12\linewidth}
            \centering
            \includegraphics[width=0.98\linewidth]{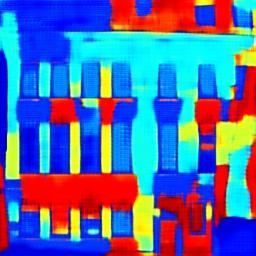}
            \includegraphics[width=0.98\linewidth]{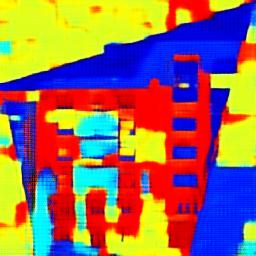}
            \includegraphics[width=0.98\linewidth]{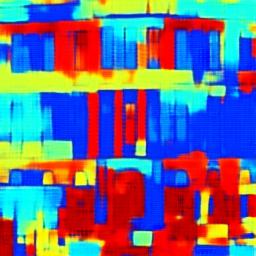}
            \stackunder[5pt]{\includegraphics[width=0.98\linewidth]{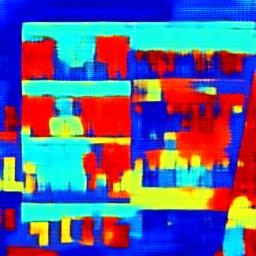}}{distorting}
        \end{subfigure}
    \caption{Qualitative results of \modelNameShort against \textsc{facade2label} under the distorting attack scheme. Similar to the results of \textsc{str2seg} (Figure \ref{fig:str2seg-start}), the color patterns are 
    disorganized. While the ``expected'' results display orderly patterns following the input facade images, the ``distorting'' results are chaotic. Red blocks (representing background) in the corners of the ``expected'' results are also often replaced with cyan blocks (representing windows) in the ``distorting'' results.}
    \label{fig:facade2label}
\end{figure*}

%% file: thefigs/app_final_visuals/night.tex
\begin{figure*}[t]
    \centering
    	\begin{subfigure}{.12\linewidth}
            \centering
            \includegraphics[width=0.98\linewidth]{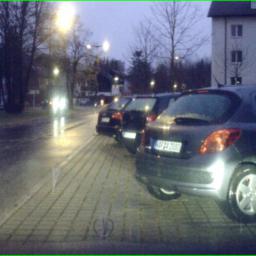}
            \includegraphics[width=0.98\linewidth]{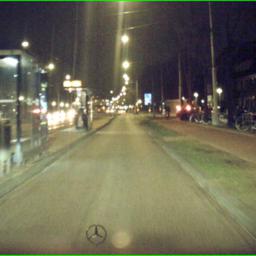}
            \includegraphics[width=0.98\linewidth]{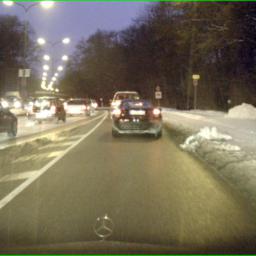}
            \stackunder[5pt]{\includegraphics[width=0.98\linewidth]{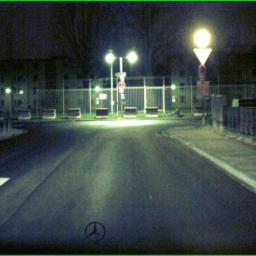}}{input}
    	\end{subfigure}
    	    \begin{subfigure}{.12\linewidth}
            \centering
            \includegraphics[width=0.98\linewidth]{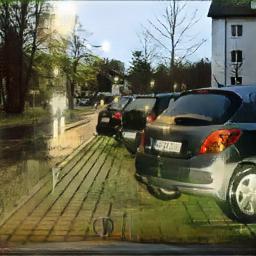}
            \includegraphics[width=0.98\linewidth]{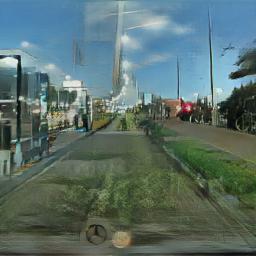}
            \includegraphics[width=0.98\linewidth]{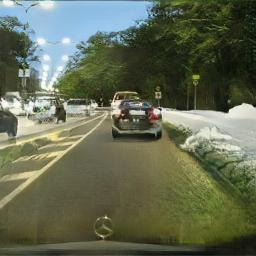}
            \stackunder[5pt]{\includegraphics[width=0.98\linewidth]{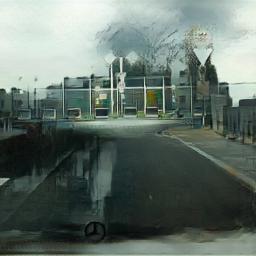}}{expected}
    	\end{subfigure}
        \begin{subfigure}{.12\linewidth}
            \centering
            \includegraphics[width=0.98\linewidth]{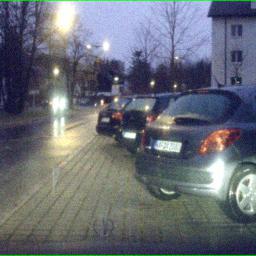}
            \includegraphics[width=0.98\linewidth]{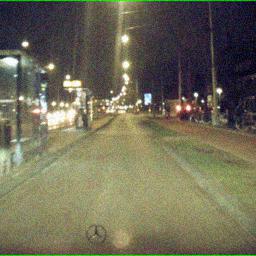}
            \includegraphics[width=0.98\linewidth]{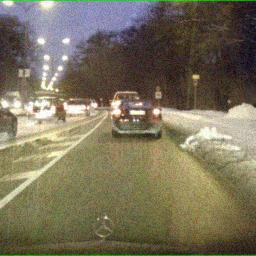}
            \stackunder[5pt]{\includegraphics[width=0.98\linewidth]{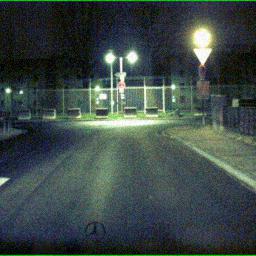}}{adversarial}
    	\end{subfigure}	
        \begin{subfigure}{.12\linewidth}
            \centering
            \includegraphics[width=0.98\linewidth]{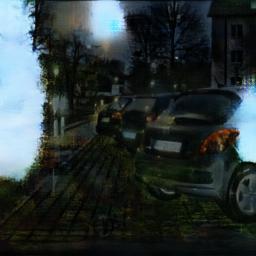}
            \includegraphics[width=0.98\linewidth]{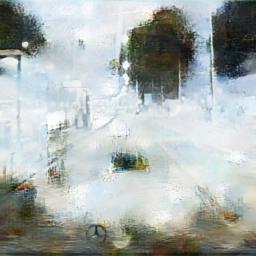}
            \includegraphics[width=0.98\linewidth]{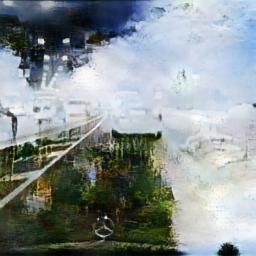}
            \stackunder[5pt]{\includegraphics[width=0.98\linewidth]{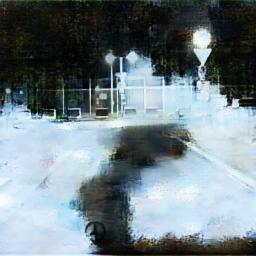}}{distorting}
        \end{subfigure}
        \begin{subfigure}{.12\linewidth}
            \centering
            \includegraphics[width=0.98\linewidth]{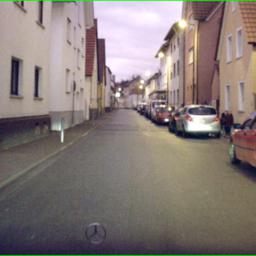}
            \includegraphics[width=0.98\linewidth]{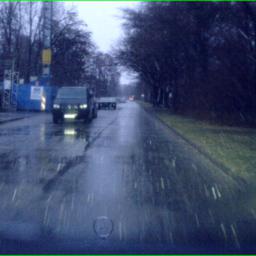}
            \includegraphics[width=0.98\linewidth]{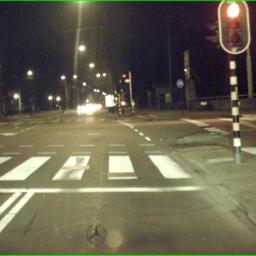}
            \stackunder[5pt]{\includegraphics[width=0.98\linewidth]{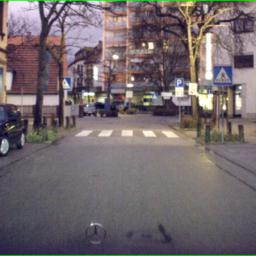}}{input}
    	\end{subfigure}
    	    \begin{subfigure}{.12\linewidth}
            \centering
            \includegraphics[width=0.98\linewidth]{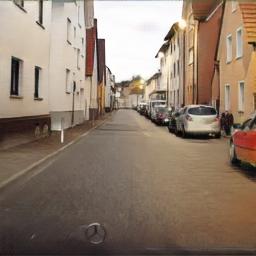}
            \includegraphics[width=0.98\linewidth]{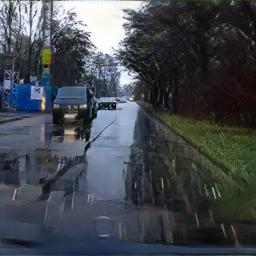}
            \includegraphics[width=0.98\linewidth]{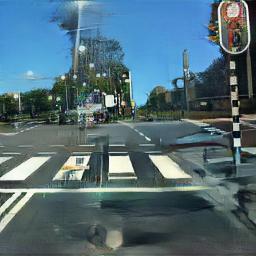}
            \stackunder[5pt]{\includegraphics[width=0.98\linewidth]{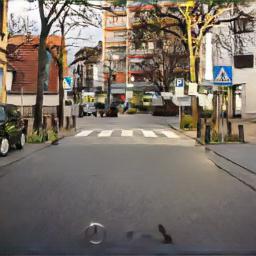}}{expected}
    	\end{subfigure}
        \begin{subfigure}{.12\linewidth}
            \centering
            \includegraphics[width=0.98\linewidth]{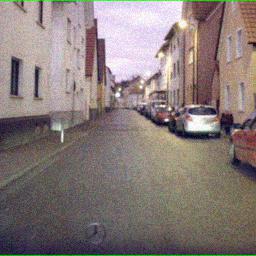}
            \includegraphics[width=0.98\linewidth]{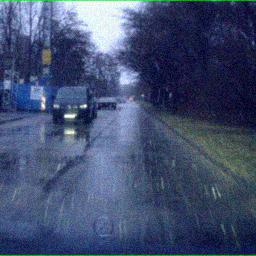}
            \includegraphics[width=0.98\linewidth]{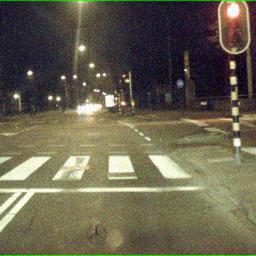}
            \stackunder[5pt]{\includegraphics[width=0.98\linewidth]{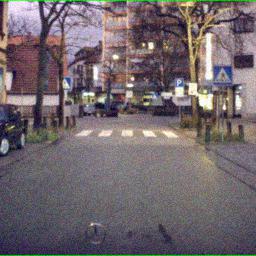}}{adversarial}
    	\end{subfigure}	
        \begin{subfigure}{.12\linewidth}
            \centering
            \includegraphics[width=0.98\linewidth]{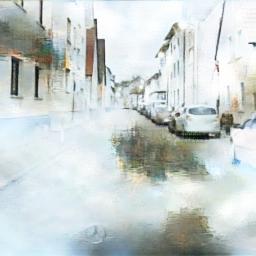}
            \includegraphics[width=0.98\linewidth]{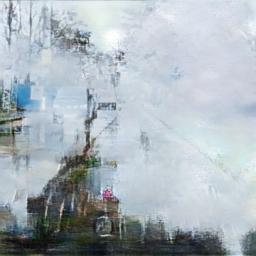}
            \includegraphics[width=0.98\linewidth]{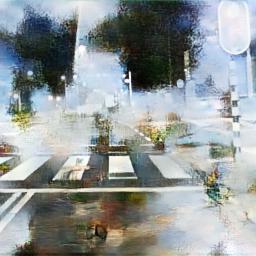}
            \stackunder[5pt]{\includegraphics[width=0.98\linewidth]{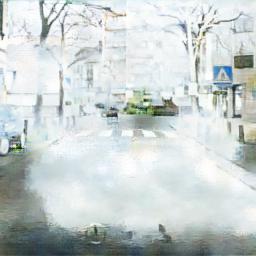}}{distorting}
        \end{subfigure}
    \caption{Qualitative results of \modelNameShort against \textsc{night2day} under the distorting attack scheme. While in the ``expected'' columns, \textsc{night2day} works as expected and produces clear images of the same street scenes as in the ``input'' columns, when given adversarial images crafted by \modelNameShort, \textsc{night2day} outputs figures obscured by black or white colors. Such adversarial attack results may hinder further processing of the street scene images.}
    \label{fig:night2day}
\end{figure*}